\documentclass[11pt,a4paper,onecolumn]{IEEEtran}

\usepackage{algorithm}
\usepackage{algorithmic}
\usepackage[colorlinks=true,citecolor=magenta,urlcolor=blue]{hyperref}
\usepackage{cite}
\usepackage{graphicx,epstopdf}
\usepackage{rotating}
\usepackage{epsfig}
\usepackage[cmex10]{amsmath}
\usepackage{amssymb}
\usepackage{amsthm}
\usepackage{float}
\usepackage[tight,footnotesize]{subfigure}
\usepackage{array}
\usepackage{color}
\usepackage{multirow}
\graphicspath{{images_final/}}
\AppendGraphicsExtensions{.tif}
\usepackage{footnote}
\makesavenoteenv{table}

\ifCLASSINFOpdf
\else
\fi

\DeclareMathOperator*{\argmax}{arg\,max}
\begin{document}

\title{Disaggregation of Remotely Sensed Soil Moisture in Heterogeneous Landscapes using Holistic Structure based Models} 
\author{Subit~Chakrabarti,~\IEEEmembership{Student~Member,~IEEE}, ~Jasmeet~Judge,~\IEEEmembership{Senior~Member,~IEEE},~Anand~Rangarajan,~\IEEEmembership{Member,~IEEE},\\~Sanjay~Ranka,~\IEEEmembership{Fellow,~IEEE}.%
\thanks{
A version of this manuscript has been submitted to the IEEE\textsuperscript{\textcopyright} Transactions on Geoscience and Remote Sensing. 

S. Chakrabarti and J. Judge are with the Center for Remote Sensing, 
Agricultural and Biological Engineering Department, 
Institute of Food and Agricultural Sciences, and the Department of Electrical Engineering, University of Florida, Gainesville, USA; A. Rangarajan and S. Ranka are with the Department of Computer \& Information Science \& Engineering, University of Florida, Gainesville.
E-mail: \href{mailto:subitc@ufl.edu}{subitc@ufl.edu}

This work was supported in part by the NASA-Terrestrial Hydrology Program (THP)-NNX13AD04G. The authors acknowledge computational resources and support provided by the University of Florida High-Performance Computing Center for all the model simulations conducted in this study. The authors thank the anonymous reviewers for their valuable suggestions and comments.
}}



\maketitle

\begin{abstract}
\boldmath
In this study, a novel machine learning algorithm is presented for disaggregation of satellite soil moisture (SM) based on self-regularized regressive models (SRRM) using high-resolution correlated information from auxiliary sources. It includes regularized clustering that assigns soft memberships to each pixel at fine-scale followed by a kernel regression that computes the value of the desired variable at all pixels. Coarse-scale remotely sensed SM were disaggregated from 10km to 1km using land cover, precipitation, land surface temperature, leaf area index, and \textit{in-situ} observations of SM. This algorithm was evaluated using multi-scale synthetic observations in NC Florida for heterogeneous agricultural land covers. It was found that the root mean square error (RMSE) for 96\% of the pixels was less than 0.02 $\mathrm{m}^3/\mathrm{m}^3$. The clusters generated represented the data well and reduced the RMSE by upto 40\% during periods of high heterogeneity in land-cover and meteorological conditions. The Kullback Leibler divergence (KLD) between the true SM and the disaggregated estimates is close to 0, for both vegetated and baresoil landcovers. The disaggregated estimates were compared to those generated by the Principle of Relevant Information (PRI) method. The RMSE for the PRI disaggregated estimates is higher than the RMSE for the SRRM on each day of the season. The KLD of the disaggregated estimates generated by the SRRM is at least four orders of magnitude lower than those for the PRI disaggregated estimates, while the computational time needed was reduced by three times. The results indicate that the SRRM can be used for disaggregating SM with complex non-linear correlations on a grid with high accuracy.
\end{abstract}
\newpage

\section{Introduction}
SM is a key governing factor in surface and sub-surface hydrological and agricultural models as it regulates land-atmosphere interactions. It has also been recognized as an \textcolor{black}{essential climate variable} by the Global Climate Observing System~\cite{Bojinski2014}. Representational models of weather\cite{Fennessy2009,Douville2000,Koster2004}, crop growth\cite{Tubiello2002}, ecosystem and carbon cycle processes\cite{Yuste2005,Friend2007}, dust generation\cite{Fecan1999}, trace gas fluxes\cite{Holtgrieve2006}, and agricultural drought\cite{Narasimhan2005,Chakrabarti2013} require soil moisture data at a fine spatial resolution. Recent satellite missions, including the European Space Agency (ESA) Soil Moisture and Ocean Salinity (SMOS) and the National Aeronautics and Space Administration (NASA) Soil Moisture Active Passive (SMAP) missions~\textcolor{black}{\cite{entekhabi2008}}, provide for SM retrievals at unprecedented spatial resolutions of tens of kilometres every 2-3 days, with worldwide coverage. However, models simulating physical processes for agricultural regions need SM at even finer scales of 1 km\cite{Chakrabarti2013}. Disaggregation addresses this discrepancy in scales by generating local fine-resolution data from coarse-resolution data obtained from satellites.

Most of the disaggregation techniques broadly fall into three approaches. The first approach is based on the assumption that spatial disaggregation follows a known hierarchical model such as fractal interpolation, power-law or temporal persistence across scales. Methods using this approach usually assume static vegetation and micro-meteorology for a given area, due to the difficulties associated with parametrizing weather and land cover data across temporal and spatial scales in such models. However, the static assumption in this approach introduces large errors in realistic applications. The second approach uses empirical models based on statistical and geo-statistical methods, such as regression, co-kriging and block kriging, and fractal interpolation. The third approach employs statistical models based on the Triangle Method\cite{Piles2009,Piles2011,Piles2014} to extrapolate the dependant data within the hypothetical triangle formed by the observed data. The robustness of the statistical methods over heterogeneous vegetation and weather conditions remain mostly untested. Treating each pixel as a sample instead of using spatial information to regularize the disaggregation results in salt and pepper noise due to spatial auto-correlation\cite{Jiang2013}. Moreover, these approaches use second order metrics, which do not leverage all the information in the data that is necessary in a highly non-linear regression problem such as disaggregation\cite{Principe2010}.

A recently implemented disaggregation algorithm\cite{Chakrabarti2014} based on the principle of relevant information (PRI) addresses the above inadequacies by utilizing the full probability density function of a set of training observations, rather than second order moments, to approximate a transformation function that relates micro-meteorological data recorded in a region to \emph{in-situ} soil moisture (SM). It uses the transformation function to generate an initial set of SM values for the rest of the data set. The disaggregated SM is obtained by iterating between the coarse scale SM values and the initial SM values using an information theoretic cost function. \textcolor{black}{ The PRI method was compared to the widely used disaggregation algorithm based on a second order regression using the Triangle Method\cite{Piles2011}. It was found to have lower disaggregation errors, especially for complex noise models added to the coarse resolution SM. Notably, the Kullback-Leibler distance between the true and disaggregated SM was 50\% lower for the PRI method, compared to the Triangle Method.} This is because methods based on the 2nd order Triangular or Quadrilateral regressions do not have separate steps for error-bias and error-variance controls and rely on the data being well-posed to achieve a balance between error-bias and error-variance. Although the PRI method results in low disaggregation errors, training a fully Bayesian transformation function is computationally intensive. Additionally, it requires a comprehensive training set for the initial estimate of the multi-dimensional PDF to converge. In this study, a self-regularized regressive model (SRRM) is used to disaggregate SM. It is expected to be less computationally intensive as it uses auxiliary features correlated to SM to perform clustering of pixels and subsequently trains a single model for each cluster. Furthermore, it requires fewer samples for training.  

\textcolor{black}{The goal of this study is to develop and implement a novel machine learning algorithm to disaggregate coarse scale remotely sensed SM using auxiliary fine-scale data. The primary objectives of this study are to - 1) develop an algorithm to identify contiguous regions of similarity in gridded images and subsequently, for each region, use kernel regression to estimate a disaggregation model for each region;} 2) implement this algorithm to estimate SM at 1 km using SM at 10 km and other spatially correlated variables in the region such as land surface temperature (LST), leaf area index (LAI), land cover (LC) and precipitation (PPT); and 3) evaluate the SRRM-based methodology and compare it with the PRI method using a synthetic dataset. 

Section~\ref{Sec:Models} describes the theoretical details of the disaggregation framework based on self-regularized regressive models and provides a brief description of the PRI algorithm for disaggregation. Section~\ref{Sec:Methodology} illustrates the steps for the implementation of the SRRM and presents the disaggregation results for SM at 1~km  and Section~\ref{Sec:Conclusion} summarizes the important results, concludes the paper, and outlines the scope for future studies.

\section{Disaggregation Framework}
\label{Sec:Models}


Disaggregation is an ill-conditioned problem that is limited physically by the convolution of the point spread function of the imaging system. This constrains the generation of fine-scale data from coarse-scale data. Additional spatially correlated information is needed to regularize the fine-scale estimates. Methods that use regression to bridge the difference in scales have to use regularization to address the multiplicity of solutions. The SRRM addresses this problem by using a clustering algorithm to create a number of regions of similarity which subsequently, are used in a kernel regression framework. This is described in more detail in the following sections. Using spatial regions or dynamic conglomerations of pixels to generate models instead of treating each pixel in a sample-based method also reduces the effect of spatial autocorrelation on the disaggregated estimates. 

\subsection{Disaggregation Framework based on Self-Regularized Regressive Models (SRRM)}
\label{Sec:OurModel}

\begin{figure}[t]
\centering\noindent
\centering\noindent\includegraphics[width=3.5 in]{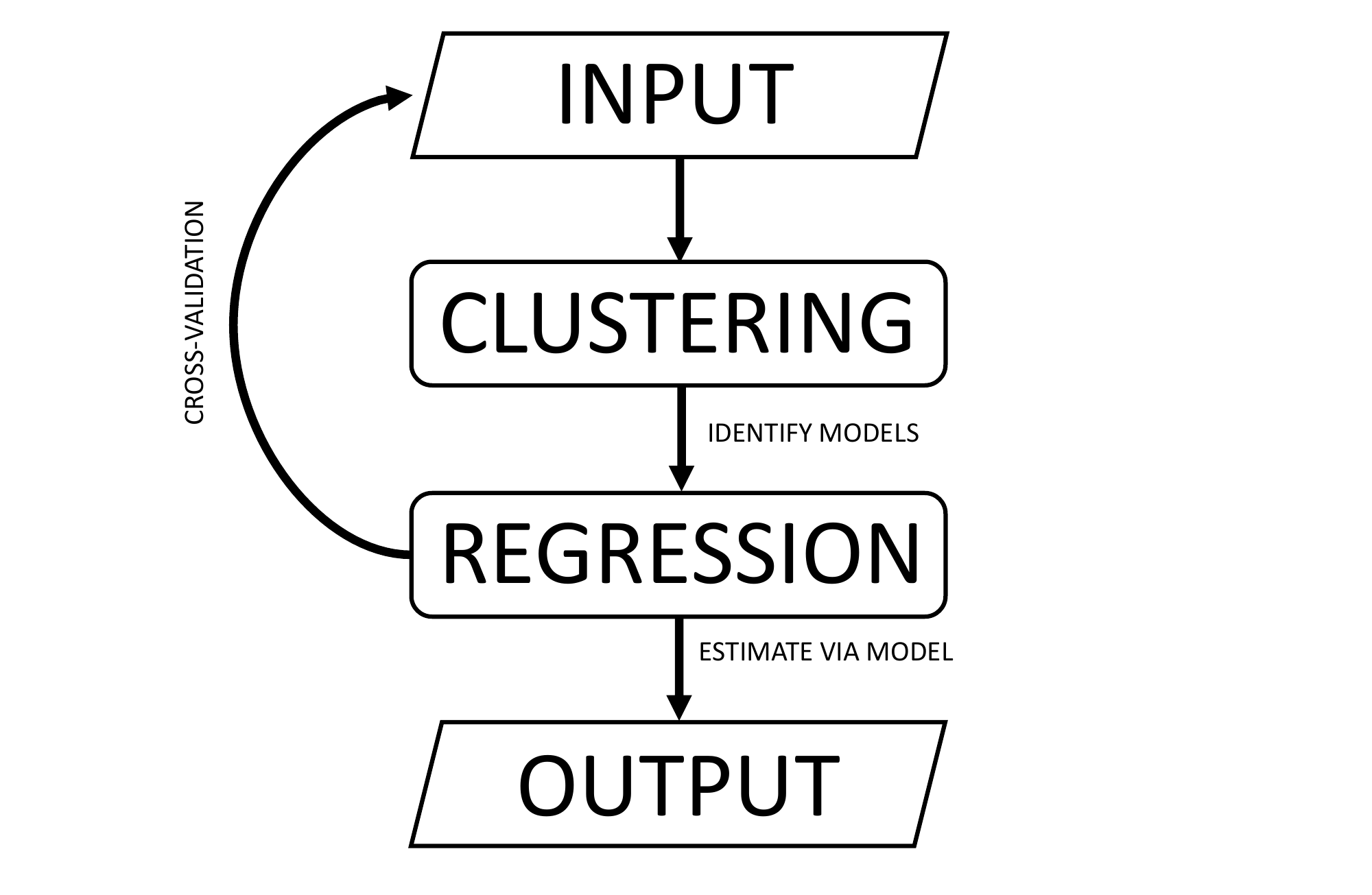}
\caption{Flowchart of the SRRM based algorithm.}
\label{fig:organization}
\end{figure}

\begin{figure}[t]
\centering\noindent
\includegraphics[width=7 in ,keepaspectratio=true]{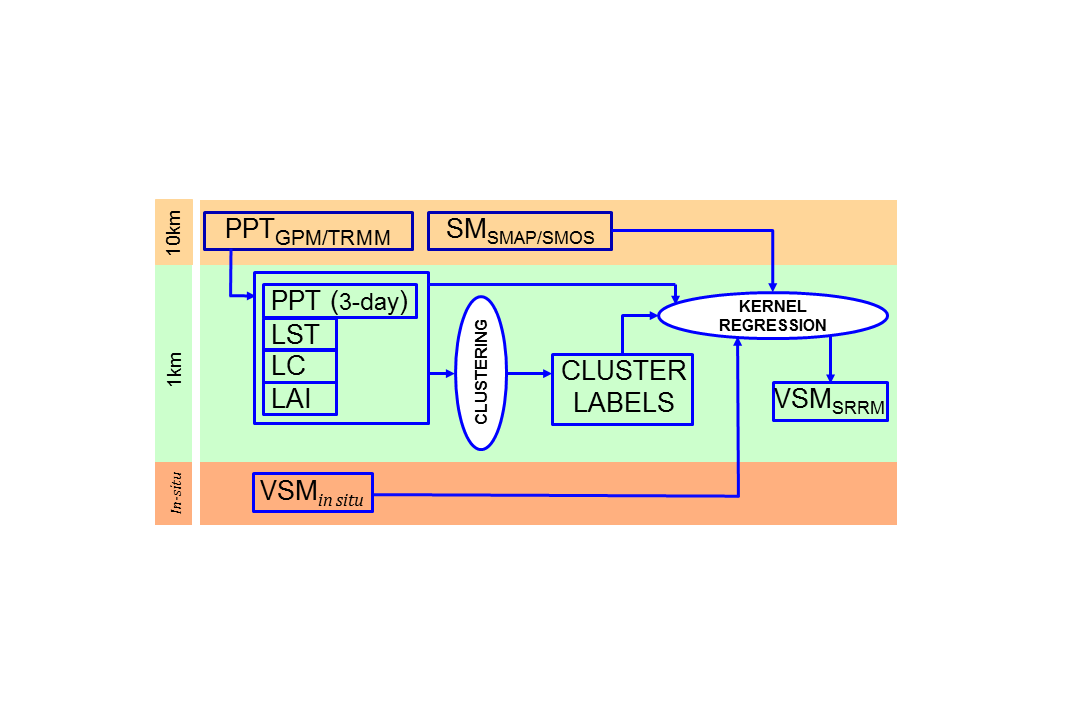}
\caption{Flow diagram of the Self-regularized Kernel Regression models.}
\label{fig:flow}
\end{figure}

\textcolor{black}{In this study, contiguous regions are identified in multi-dimensional correlated data using clustering and subsequently a regression model is trained for each cluster for disaggregation.} The membership vector of every pixel to a region, and thus to a model, is soft and constrained to sum to one across the space of models. The models themselves are trained using a kernel regression based method. It is a novel way to account for correlated features using algorithms that require an IID (independence and identical distribution) assumption\cite{Jiang2013}. Figure\footnote{All \hyperref[Figures]{Figures} and \hyperref[Tables]{Tables} are included at the end of the manuscript for clarity.}\addtocounter{footnote}{-1}\addtocounter{Hfootnote}{-1}~\ref{fig:organization} shows a flow diagram of the algorithm for generating disaggregated estimates. The overall organization and the datasets involved is shown in Figure~\ref{fig:flow}. The two steps of the algorithm include clustering and kernel regression, as follows. 

\subsubsection{Information theoretic (IT) clustering based on the Cauchy-Schwarz Distance}
\begin{figure}[t]
\centering\noindent
\includegraphics[width=7 in ,keepaspectratio=true]{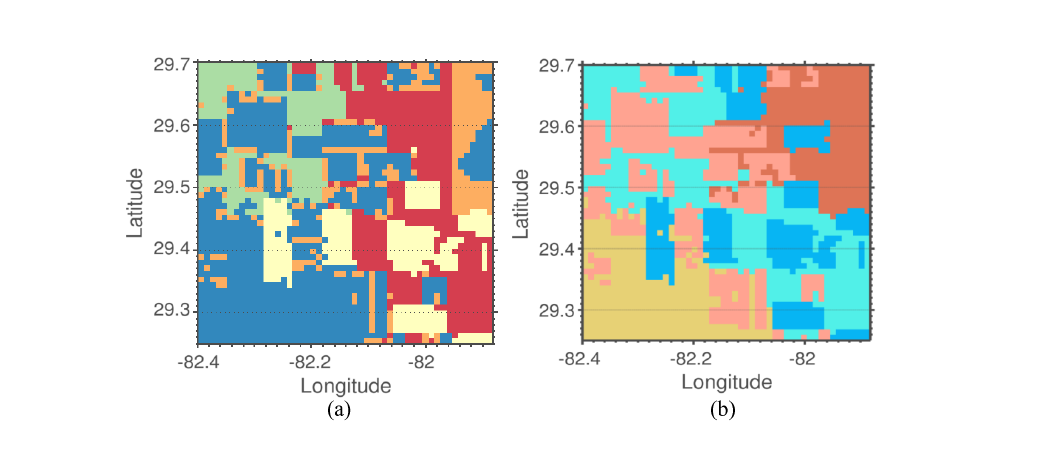}
\caption{Clustering result obtained from (a) DCS based clustering algorithm, and (b) K-Means clustering algorithm.}
\label{fig:clustering}
\end{figure}

\textcolor{black}{Commonly used clustering methods, such as the K-Means~\cite{likas2003}, assume hyper-spherical or hyper-elliptical clusters~\cite{jenssen2005}. With gridded remotely sensed data, prior assumptions about cluster shapes are not advisable and lead to noise in the clustering result, as shown in Figure~\ref{fig:clustering}(a). Instead, in the IT clustering method the generalized proximity regions are identified using a regularized variant of a clustering method based on information theory\cite{jenssen2005}. The clusters are constructed using the probability density functions~(PDFs) of the data, resulting in clusters that are representative of the input data, as shown in Figure~\ref{fig:clustering}(b).} For any two vectors $\textbf{x}$ and $\textbf{y}$, the Cauchy-Schwarz inequality is,
\begin{equation}
\label{eq:CS}
-\mathrm{log}\left(\frac{|<\textbf{x},\textbf{y}>|}{\sqrt{\|\textbf{x}\|^2  \|\textbf{y}\|^2}}\right)\geq 0
\end{equation}
where $<\textbf{x},\textbf{y}>$ is the inner product of vectors $\textbf{x}$ and $\textbf{y}$. For PDFs $p(x)$ and $q(x)$, the inner product is defined as, $<p,q> = \int p(x)q(x) \mathrm{d}x$ over the support for the distributions $p$ and $q$. Then, the Cauchy-Schwarz inequality in a metric space spanned by the PDF is,

\begin{equation}
\label{eq:JCS}
-\mathrm{log}\left(\frac{|\int p(x)q(x)\mathrm{d}x|}{\sqrt{\int p^2(x)\mathrm{d}x \int q^2(x) \mathrm{d}x}}\right)\geq 0
\end{equation} 
If $p(x)$ is calculated using pixels lying in cluster $C_1$ and $q(x)$ is calculated using pixels lying in cluster $C_2$, the maximum separation is obtained between clusters when the left-hand side of Equation~\ref{eq:JCS}, the Cauchy Schwarz distance ($D_{CS}$), is maximized. Since logarithm is a monotonically increasing function, only the argument of the logarithm in $D_{CS} = - \mathrm{log} J_{CS}(p,q)$ can be equivalently minimized using gradient descent based optimization. An estimator $\hat{J}_{CS}$ of $J_{CS}(p,q)$ can be constructed from data-samples and extended to the case of multiple clusters by using a membership vector.
\begin{equation}
\label{eq:JCS_estimator}
\hat{J}_{CS}(\mathbf{m}_1,\dots,\mathbf{m}_N) = \frac{\frac{1}{2} \sum_{i=1}^{N} \sum_{j=1}^{N} \left(1 - \textbf{m}_i^\mathrm{T} \textbf{m}_j\right) G_{\sigma \sqrt{2}} \left(\mathbf{x}_i,\mathbf{x}_j\right)}{\sqrt{\prod_{k=1}^{K} \sum_{i=1}^{N} \sum_{j=1}^{N} m_{ik}m_{jk}G_{\sigma \sqrt{2}}(\mathbf{x}_i,\mathbf{x}_j)}}
\end{equation}
where $\mathbf{m}_i$ is a soft K-dimensional vector, where the $k^{th}$ element expresses the degree of membership to the $k^{th}$ cluster. $K$ is the total number of clusters which has to be supplied as input. $G_{\sigma\sqrt{2}}(\cdot,\cdot)$ is derived from convolution of two Gaussian kernels, defined as $G_{\sigma\sqrt{2}}(\mathbf{x}_i,\mathbf{x}_j) =\mathrm{exp}\left(-\frac{\| \mathbf{x}_i-\mathbf{x}_j \|_2^2 }{2\sigma^2}\right)$. A regularized version can be used as an objective function of clustering, 
\begin{equation}
\label{eq:JCS_estimator_reg}
\hat{J}_{CS}^{REG}(\mathbf{m}_1,\dots,\mathbf{m}_N) = \frac{\frac{1}{2} \sum_{i=1}^{N} \sum_{j=1}^{N} \left(1 - \textbf{m}_i^\mathrm{T} \textbf{m}_j\right) G_{\sigma \sqrt{2}} \left(\mathbf{x}_i,\mathbf{x}_j\right)}{\sqrt{\prod_{k=1}^{K} \sum_{i=1}^{N} \sum_{j=1}^{N} m_{ik}m_{jk}G_{\sigma \sqrt{2}}(\mathbf{x}_i,\mathbf{x}_j)}} - \psi \sum_{i=1}^{N} \sum_{k=1}^{K} m_{ik} \mathrm{log}\left(m_{ik}\right)
\end{equation}
The second term of the objective function is an estimate of the Shannon Entropy of the membership vectors and serves to regularize the membership vectors such that the model selection is sufficiently sparse.  Getting the correct membership vector then is equivalent to solving this constrained optimization problem:
\begin{equation}
\label{eq:optimization}
\text{min}_{\substack{\mathbf{m}_1,\dots,\mathbf{m}_N}}\hat{J}_{CS}^{REG}(\mathbf{m}_1,\dots,\mathbf{m}_N) \quad \text{subject to } \mathbf{m}_j^\mathrm{T}\mathbf{1}_{K\times 1} - 1 = 0, \quad j = 1,\dots ,N 
\end{equation}
where $\mathbf{1}_{K\times 1}$ is a $K\times 1$ vector whose elements are all one. Consider $m_{ik} = \textit{v}_{ik}^2,k=1,\dots,K$ which corresponds to a form that can be optimized by using Lagrange multipliers. The Lagrangian can be expressed as,
\begin{equation}
\label{eq:Lagrangian}
L = \hat{J}_{CS}^{REG}(\mathbf{v}_1,\mathbf{v}_2,\ldots,\mathbf{v}_N) + \sum_{i=1}^N \lambda_i (\mathbf{v}_i^\mathrm{T} \mathbf{v}_i - 1)
\end{equation}
The optimization problem Equation~\ref{eq:Lagrangian} amounts to adjusting vectors $\mathbf{v}_i, i=1,\dots,N$ such that,
\begin{equation}
\label{eq:gradient}
\frac{\partial \hat{J}_{CS}^{REG}}{\partial \mathbf{v}_i} = \left(\frac{\partial \hat{J}_{CS}^{REG}}{\partial \mathbf{m}_i}^\mathrm{T} \frac{\partial \mathbf{m}_i}{\partial \mathbf{v}_i}\right)^\mathrm{T} = \Gamma \frac{\partial \hat{J}_{CS}^{REG}}{\partial \mathbf{m}_i} \to 0,
\end{equation}
where $\Gamma = \mathrm{diag}(2\sqrt{m_{i1}},\dots,2\sqrt{m_{iK}})$ is the magnitude normalizing factor. \textcolor{black}{The memberships are forced to be positive by adding a constant of small magnitude, $\alpha \sim 0.05$ to all elements of $\Gamma$.} The Lagrange Multipliers then, after constructing the necessary Lagrange Function is given by 
\begin{equation}
\label{eq:lambda}
\lambda_i = \frac{1}{2} \sqrt{\frac{\partial \hat{J}_{CS}^{REG}}{\partial \mathbf{v}_i}^\mathrm{T}\frac{\partial \hat{J}_{CS}^{REG}}{\partial \mathbf{v}_i}}
\end{equation}
The updated vector for the next iteration is,
\begin{equation}
\label{eq:membership}
\mathbf{v}_{i}^+ = -\frac{1}{2\lambda _i}\frac{\partial \hat{J}_{CS}^{REG}}{\partial \mathbf{v}_i}
\end{equation}
The square of the membership vectors are initialized as $\mathbf{v}_i = |\mathcal{N}(0;\gamma^2 \mathbf{I})|$, where $\mathcal{N}$ denotes the Gaussian distribution and $\gamma$ is a very small number.

\subsubsection*{Stochastic Approximation of the Gradient and Computational Complexity}
\textcolor{black}{
If $\hat{J}_{CS}$ is represented as $\frac{U}{V}$, then the gradient of $\hat{J}_{CS}^{REG}$ can be calculated as:
\begin{align}
\label{eq:Lagrange}
&\frac{\partial \hat{J}_{CS}^{REG}}{\partial \mathbf{m}_i} = \frac{V \frac{\delta U}{\partial \mathbf{m}_i} - U\frac{\partial V}{\partial \mathbf{m}_i}}{V^2} - \psi \sum_{k=1}^K \left( 1 + \mathrm{log } (m_{ik}) \right)
\end{align}
where U and V are defined as,
\begin{align}
&U = \frac{1}{2} \sum_{i=1}^{N}\sum_{j=1}^{N} \left(1 - \textbf{m}_i^T \textbf{m}_j\right) G_{\sqrt{2}\sigma}(\mathbf{x}_i,\mathbf{x}_j)\quad
\text{and} \quad  V = \sqrt{\prod_{k=1}^K v_k} 
\end{align}
and the gradients of U and V are defined as,
\begin{align}
\frac{\partial U}{\partial \mathbf{m}_j} = -\sum_{j=1}^N \mathbf{m}_j G_{\sigma \sqrt{2}} (\mathbf{x}_i,\mathbf{x}_j) \quad
\text{and} \quad\frac{\partial V}{\partial \mathbf{m}_i} = \frac{1}{2} \sum_{{k'}=1}^K \sqrt{\frac{\prod_{\substack{k = 1 \\ {k}\neq k'}}^K v_{k}} {v_{k'}}} \frac{\partial v_{k'}}{\partial \mathbf{m}_i},
\end{align}
where $v_k = \sum_{i=1}^N \sum_{j=1}^N \mathbf{m}_{ik} \mathbf{m}_{jk} G_{\sqrt{2}\sigma}(\mathbf{x}_i,\mathbf{x}_j)$ \\ and $\frac{\partial v_{k'}}{\partial m_i} = \left[0,\dots, 2\sum_{j=1}^N \mathbf{m}_j(k')G_{\sqrt{2}\sigma}(\mathbf{x}_i,\mathbf{x}_j),\dots,0\right]^T$.\\ 
}
\subsubsection*{Kernel Annealing}
The objective function in Equation~\ref{eq:JCS_estimator_reg} has local minima that can inhibit the performance of this algorithm. To ensure that the clustering solution is globally, and not just locally, minimum the kernel width is gradually decreased in this algorithm over the course of iterations. The initial value of the kernel is chosen according to the \textcolor{black}{Silverman's rule of thumb\cite{Silverman1986}} given by
\begin{equation}
\label{eq:Silverman}
\sigma_{\mathrm{SIL}} = \sigma_{X} \left( 4N^{-1}\left(2d+1\right)^{-1}\right)^{\frac{1}{d+4}}
\end{equation}
where $d$ is the dimensionality of the data, $N$ is the number of samples and $\sigma^2_{X} = d^{-1} \sum_i \sum_{X_{ii}}$ and $\sum_{X_{ii}}$ is the diagonal values of the sample covariance matrix. The lower value of the kernel size is set to $\sigma_{\mathrm{LOW}}=\frac{\sigma_{\mathrm{SIL}}}{4}$. Thus the annealing rate is,
\begin{equation}
\label{eq:AnnealingRate}
r = \frac{\sigma_{\mathrm{SIL}}-\sigma_{\mathrm{LOW}}}{N_{\mathrm{TOT}}} = \frac{3\sigma_{\mathrm{SIL}
}}{4N_{\mathrm{TOT}}}
\end{equation}

\subsubsection{Regularized Kernel Regression}

A kernel based regression technique that uses a training set of pixels and fits a function to it, by minimizing the representational error, is used to generate the disaggregated estimates. Ridge regression\textcolor{black}{\cite{Kibria2003}} is a parametric regression technique that adds a scaled regularizing term to the cost function. \textcolor{black}{This improves the stability of the regression as the added $\mathrm{L}_2$ normed term in the cost function results in smaller eigenvalues.} The cost function is, 

\begin{equation}
\mathcal{E}\left( \mathbf{w},\mathbf{x}\right)   = \frac{1}{2} \sum_i (y_i - w^{\mathbf{T}}\mathbf{x}_i)^2 + \frac{1}{2} \mu \|\mathbf{w}\|^2
\end{equation}

The weights can be calculated by differentiating the error cost function with respect to the weights and setting it to zero. 

\begin{equation}
\frac{\partial\mathcal{E}}{\partial \mathbf{w}} = 0 \implies \mathbf{w} = \left(\sum_i \mathbf{x}_i \mathbf{x}_i^\mathbf{T} + \mu \mathbf{I}\right)^{-1} \left(\sum_i y_i \mathbf{x}_j\right)
\end{equation}

For computation in a Reproducing Kernel Hilbert Space (RKHS), then the inner-products can be replaced with a kernel evaluation. Let $\mathcal{H}$ be a Hilbert space with an inner-product metric $<\cdot,\cdot>_{\mathcal{H}}$. Then according to the representer theorem, a kernel function $\kappa(\mathbf{x},\mathbf{y})$ exists on $\mathbb{R}^\mathrm{N} \times \mathbb{R}^\mathrm{N}$ such that $<\mathbf{x},\mathbf{y}>_{\mathcal{H}} = \kappa(\mathbf{x},\mathbf{y})$. Now, if $\Phi:\mathbb{R}^\mathrm{N} \rightarrow \mathbb{R}^\mathrm{N}$ is a mapping that transforms the feature vector in the original vector space to $\mathcal{H}$, then the weights can de redefined as,

\begin{equation}
\mathbf{w} = (\mu \mathbf{I}_D + \Phi \Phi^\mathrm{T})^{-1} \Phi\mathbf{y}
\end{equation}
Where $D$ is the dimension of the feature space. The dimension of the feature space is not well-defined in many cases, so the weights can be rewritten using the identity, $(A^{-1}+B^\mathrm{T}C^{-1}B)^{-1}B^\mathrm{T}C^{-1} = AB^\mathrm{T}(BAB^\mathrm{T} + C)^{-1}$, 

\begin{equation}
\label{eq:weights}
\mathbf{w} = \Phi(\mu \mathbf{I}_N + \Phi^\mathrm{T}\Phi )^{-1}\mathbf{y}
\end{equation}
The weight vector $\mathbf{w}$ can be calculated using a training set of observations where $\mathbf{y}$ is known. This can then be used to calculate the estimated value for a new data-point $\mathbf{x'}$, 
\begin{align}
\label{eq:regression}
\hat{y} &= \mathbf{w}^\mathrm{T} \Phi(\mathbf{x}') \\ \nonumber
&= \mathbf{y}(\mu \mathbf{I}_N + \Phi^\mathrm{T}\Phi )^{-1}\Phi^{\mathrm{T}}\Phi(\mathbf(x')) \\ \nonumber
&= \underbrace{\mathbf{y}(\mu \mathbf{I}_N + \mathbf{K} )^{-1}}_{\mathbf{w}} \kappa(\mathbf{x},\mathbf{x}')
\end{align}
where $\mathbf{\mathrm{K}}$ is the Gram matrix of inner products of all the training data points. This does not address the constant that must be present in the regression. To solve this problem, the feature vector is augmented by adding a constant feature 1 to all samples.

\subsubsection{Algorithm Summary and Computational Complexity}

\begin{algorithm}                      
\caption{Disaggregation using Self-Regularized Regressive Models}          
\label{alg1}                           
\begin{algorithmic}
\REQUIRE Initialize membership vectors, $\mathbf{v}_i\leftarrow |\mathcal{N}(0;\gamma ^2\mathbf{I})|$ and number of clusters, $\mathrm{N}$ for each day of the data-set. $\mathrm{N_{DAYS}}$ is the total number of days.
\FOR{$i=0$ to $\mathrm{N_{DAYS}}$}
	\STATE \textit{Step 1: Clustering}
	\FOR{$i=1$ to $30$} 
	\STATE Calculate $\hat{J}_{CS}^{REG}$ and $\frac{\partial \hat{J}_{CS}^{REG}}{\partial \mathbf{m}_i}$ according to Equation~\ref{eq:JCS_estimator_reg} and~\ref{eq:Lagrange}.
	\STATE Update $\lambda_i$ and $\mathbf{v}_i^+$ according to Equation~\ref{eq:lambda} and~\ref{eq:membership}.
	\ENDFOR
	\STATE \textit{Step 2: Kernel Regression}
	\STATE Calculate $\mathbf{w}$ according to Equation~\ref{eq:weights} using the training set.
	\STATE Estimate the disaggregated observations, $\mathbf{\hat{y}}$ for the test set using Equation~\ref{eq:regression}.
	\STATE Run 10-fold cross-validation for the values of $\mathrm{N}$ and the cross-validation constants $\psi$ and $\mu$.
\ENDFOR 
\end{algorithmic}
\end{algorithm}

The SRRM disaggregation is summarized and shown in Algorithm~\ref{alg1}. \textcolor{black}{A ten-fold cross-validation was used to determine the number of clusters ($\mathrm{N}$) and the kernel size for the clustering ($\psi$) and the regularization weight for the regression ($\mu$). The performance of the algorithm was less sensitive to the kernel size for regression than the other parameters and was set to the standard deviation of $\mathbf{y}$ at coarse scale. }

The complexity of the $\mathrm{D}_{\mathrm{CS}}$ based clustering algorithm is $\mathcal{O}(N^2)$ for each iteration. For good convergence, ~30 iterations are needed. This is much lower than the dimensionality of the data-set and does not affect the complexity of the algorithm. To reduce the computational load, a stochastic sampling method is used. For this, the gradient is approximated by using $M$ samples out of all $N$. The complexity then becomes $\mathcal{O}(MN)$ $(M<<N)$ per iteration. $M$ can be much lesser than $N$ and the results are comparable to the original method, taking a fraction of the time. The average complexity of the ridge-regression method is~ $\mathcal{O}(N^3)$\cite{Schlkopf2001}.

\subsection{The PRI Framework}
\label{Sec:PRI}

The disaggregation methodology using PRI includes a transformation process to obtain a probabilistic relationship between the variable to be disaggregated, $\mathbf{y}$, at 1 km using auxiliary information, $\mathbf{X}$, at the same scale. A discrete formulation of the Bayes rule is used to estimate $\mathbf{y}_{\mathrm{initial}}$ at fine resolution, as given in equation (\ref{eq:bayes}), wherein $\mathbf{y}_{\mathrm{train}}^{i}$ is discretized into $k$ classes, $i\in[1,k]$, and $\mathbf{x}_{j,\mathrm{train}}^{i_1}$ is discretized into $k_j$ classes in $i_1\in[1,k_j]$, where $j$ indexes the individual variables that comprise $\mathbf{X}$, $m$.

\begin{align}
p(\mathbf{y}_{\mathrm{initial}}^{i_1}|\mathbf{X}_{\mathrm{train}}^{i_1}) & =  \frac{p(\mathbf{X}_{\mathrm{train}}^{i_1}|\mathbf{y}_{\mathrm{train}}^{i})p(\mathbf{y}_{\mathrm{train}}^{i})}{p(\mathbf{X}_{\mathrm{train}}^{i_1})} \nonumber \\
\mathbf{y}_{\mathrm{initial}}^{i} & =  \argmax_{\mathbf{y}_{\mathrm{train}}^{i}}\frac{p(\mathbf{X}_{\mathrm{train}}^{i_1}) p(\mathbf{y}_{\mathrm{train}}^{i})}{p(\mathbf{X}_{\mathrm{train}}^{i_1})} \nonumber \\
p(\mathbf{X}_{\mathrm{train}}^{i_1}) & =  \sum_{i=1}^{k}p(\mathbf{X}_{\mathrm{train}}^{i_1}|\mathbf{y}_{\mathrm{train}}^{i})p(\mathbf{y}_{\mathrm{train}}^{i})
\label{eq:bayes}
\end{align}

In the second step, $\mathbf{y}_{\mathrm{initial}}$ is merged with the observations at the coarser resolutions,  $\mathbf{y}_{\mathrm{coarse}}$ to obtain improved estimates at fine resolution,

\begin{equation}
\argmax_{\mathbf{m}}L(\mathbf{m}) = H(\mathbf{m})+\beta KL(p_{\mathbf{m}}||p_{\mathbf{y}_{\mathrm{initial}}})
\label{eq:pri}
\end{equation}
where $I(\mathbf{m})$ is the cost function, $p_{\mathbf{y}_{\mathrm{INITIAL}}}$ is the PDF of the original data, and $p_{\mathbf{m}}$ is the PDF at each iteration. $I(\mathbf{m})$ is the entropy, and $KL$ is the KL divergence. $\mathbf{m}$ is initialized to $\mathbf{y}_{\mathrm{COARSE}}$ at the first iteration. The $\beta$ is a user-defined weighting parameter that balances the redundancy and information preservation in $I(\mathbf{m})$. As the value of $\beta$ increases, the cost function gives more emphasis to KL, thus preserving more information about the data at the cost of extremely high redundancy reduction. In this study, an intermediate value of $\beta=2$ was chosen so that the PRI-image would approximate the mean level of $\mathbf{y}$ at coarse scales but will also embed the level of detail provided by the initial estimates of $\mathbf{y}$ at 1 km, to obtain morphed estimates of $\mathbf{y}$ at 1 km. \textcolor{black}{The computational complexity of the PRI algorithm is given as  $\mathcal{O}(N^3\prod k_j)$  where $k_j$ are the number of bins used to estimate the PDF of the features for the transformation function.} A detailed description of the PRI algorithm can be found in \cite{Chakrabarti2014}.

\section{Experimental Description and Results}
\label{Sec:Methodology}

\subsection{Multiscale synthetic dataset}

\begin{figure}[t]
\centering\noindent
\includegraphics[width=3.5 in ,keepaspectratio=true]{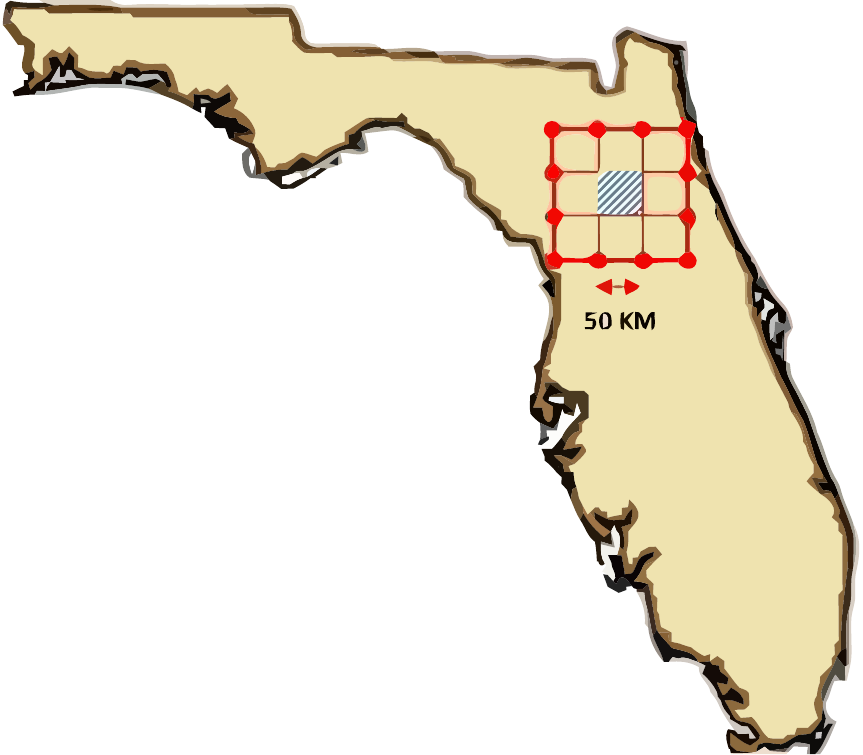}
\caption{Study region in North Central Florida. LSP-DSSAT-MB simulations were performed over the shaded $50\times50$ km$^2$ region.}
\label{fig:studysite}
\end{figure}

The proposed algorithm for disaggregation was tested using data generated by a simulation framework consisting of the Land Surface Process (LSP) model and the Decision Support System for Agrotechnology Transfer (DSSAT) model, described in \cite{Nagarajan2012}. A $50\times50$ km$^2$ region, equivalent to approximately 25 SMAP pixels at 9~km spatial resolution, was chosen in North Central Florida (see Figure~\ref{fig:studysite}) for the simulations. The region encompassed the UF/IFAS (University of Florida's Institute of Food and Agricultural Sciences) Plant Science Research and Education Unit, Citra, FL, where a series of season-long field experiments, called the Microwave, Water and Energy Balance Experiments (MicroWEXs), have been conducted for various agricultural land covers over the last decade \cite{Bongiovanni2009,Casanova2006,Lin2004}.  Simulated observations of LST $\&$ LAI were generated at 200 m for a period of one year, from January 1, 2007 through December 31, 2007. Topographic features, such as slope, were not considered in this study because the region is typically characterized by flat and smooth terrains with no run-off due to soils with high sand content. The soil properties were assumed constant over the study region. 

Fifteen-minute observations of precipitation, relative humidity, air temperature, downwelling solar radiation, and wind speed were obtained from eight Florida Automated Weather Network (FAWN) stations \cite{Fawn} located within the study region (see Figure~\ref{fig:studysite}). The observations were spatially interpolated using splines to generate the meteorological forcings at 200 m. Long-wave radiation was estimated following Brutsaert \cite{Brutsaert1975}.

\begin{figure}
\centering\noindent\includegraphics[width=7 in ,keepaspectratio=true,clip]{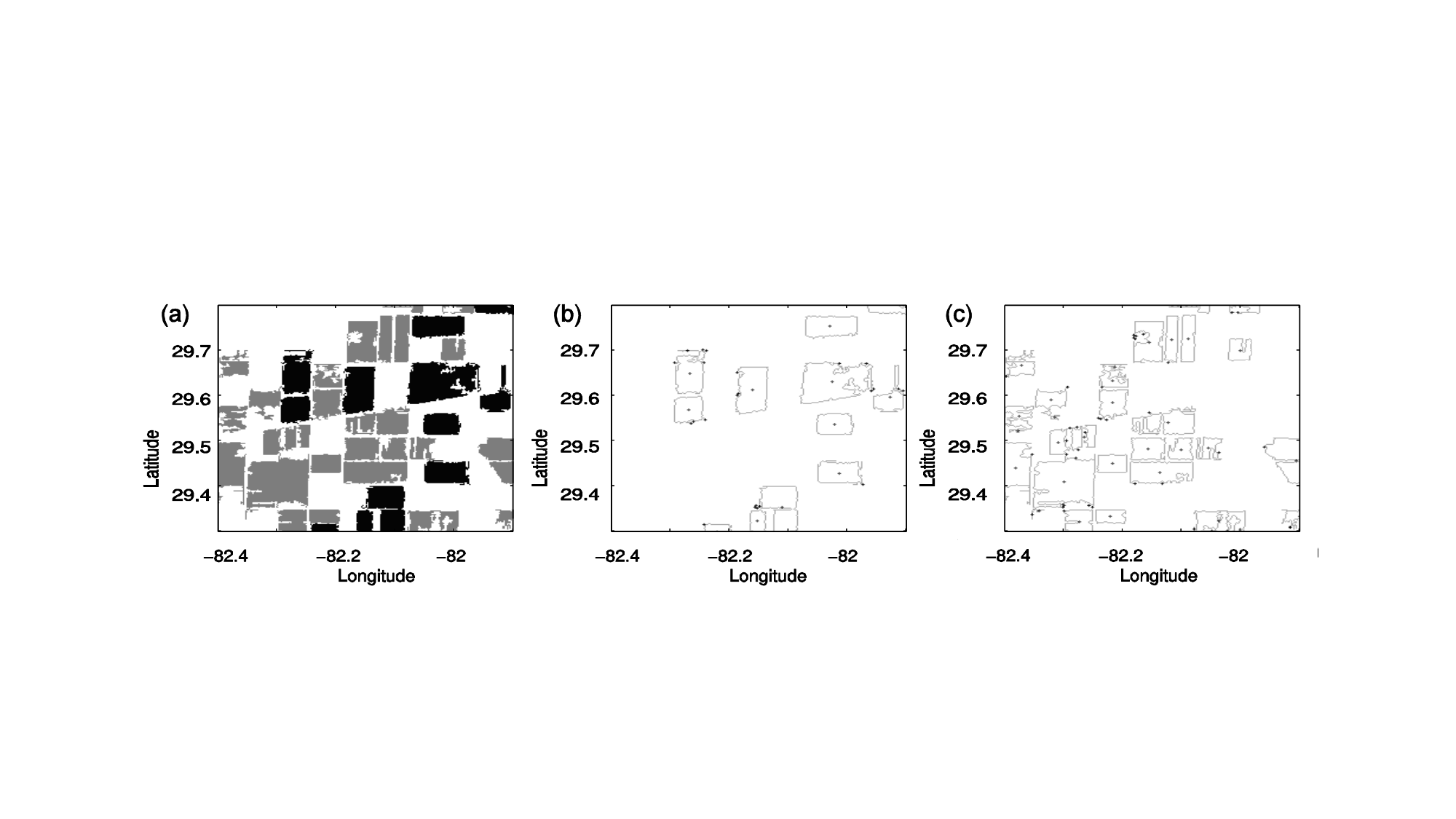}
\caption{(a) Land cover at 200m during cotton and corn seasons. White, gray, and black shades represent 
baresoil, cotton, and sweet-corn regions, respectively. Homogeneous crop fields along with centers for (b) sweet-corn and (c) cotton.}
\label{fig:landcover}
\end{figure}

The model simulations were performed over each agricultural field rather than all the pixels, to reduce computation time. 
Based upon land cover information at 200 m, contiguous, homogeneous regions of sweet-corn and cotton were identified, as shown in Figure~\ref{fig:landcover}. A realization of the LSP-DSSAT model was used to simulate LST, LAI, and PPT at the centroid of each homogeneous region, using the corresponding crop module within DSSAT. The model simulations were performed using the 200 m forcings at the centroid, as shown in Figure~\ref{fig:landcover}. Linear averaging is typically sufficient to illustrate the effects of resolution degradation \cite{Crow2003}. The model simulations at 200 m were spatially averaged to obtain PPT, LST, LAI, SM, and T$_{\textrm{B}}$ at 1 and 10 km. \textcolor{black}{The SM obtained at 1 km was divided into the training and test sets that were used as truth to evaluate the disaggregation methodology and serve as simulated ``\textit{in-situ}" measurements to train the algorithm respectively.} \textcolor{black}{ PPT, LAI and LST are typically chosen due to their high correlations with SM~\cite{Piles2011,merlin2013,Chakrabarti2014}. Other geophysical descriptors such as slope and soil texture were not used in this study because of their limited utility in a flat and primarily sandy region, such as that in North Central Florida. To simulate rainfed systems, all the water input from both precipitation and irrigation were combined together, and the ``PPT" in this study represents these combined values.}


\subsection{Disaggregation Framework based on SRRM}

The simulation period, from Jan 1 (DoY 1) to Dec 31 (DoY 365), 2007, consisted of two growing seasons of sweet corn and one season of cotton, as shown in Table\footnotemark~\ref{tab:crops}. The LST, PPT, and LAI observations at 1 km were obtained by adding \textcolor{black}{white Gaussian noise} to account for satellite observation errors, instrument measurement errors, and micro-meteorological variability, following \cite{Huang2008b,Privette2002,Crow2002}. Errors with zero mean and standard deviations of 5K, 1 mm/hour, 0.03 $\mathrm{m^3/m^3}$ and 0.1 for LST, PPT, SM  and LAI, respectively, were added to the values at 10 km.

The SRRM uses LST, 3-day PPT, LAI, LC at 1 km and SM at 10 km every 3 days as input. In the first step, the information-theoretic cost function described in Section~\ref{Sec:OurModel} is used for clustering using the inputs at 1~km and the x and y coordinates of each pixel scaled to a range of 0 and 1. This step of the algorithm uses two parameters - the number of clusters, $\mathrm{N}$ and a regularization constant, $\mu$. Both the number of clusters and the regularization constant is determined by cross-validating against the absolute mean error in SM at the end of the second step for each day. 

\begin{figure}
\centering\noindent\includegraphics[width=3.5 in]{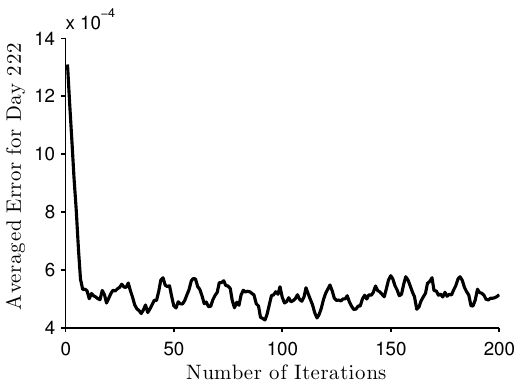}
\caption{Root mean Square error in disaggregated soil moisture at 1 km versus number of iterations of the $D_{CS}$ clustering algorithm. }
\label{fig:IterError}
\end{figure}

The optimal number of iterations that produced a usable clustering result was determined by the minimum root mean square error (RMSE) for a day when both the land cover and micro-meteorological conditions were heterogeneous, DoY 222, providing the worst case-scenario for convergence of the clustering algorithm. At the end of this step, each pixel has a vector of $\mathrm{N}$ numbers, $(m_1,m_2,\dots,m_N)$ that sum to 1 describing its membership to each of the $\mathrm{N}$ clusters. Figure~\ref{fig:IterError} shows the spatially averaged RMSE between disaggregated SM and the observations at 1 km on DoY 222 for different iterations of the clustering algorithms.  All parameters, except the number of clusters, were cross-validated for each individual iteration.The number of clusters were cross-validated once, using 50 iterations of the clustering algorithm. \textcolor{black}{For the cross validation, the training set was randomly divided into 10 equal parts. Nine parts were used for training and one part was used for evaluating the algorithm. This methodology, known as 10-fold cross-validation, is repeated ten times with different randomly selected partitions to approximate the average errors that the SRRM would incur.} The error oscillates with a mean amplitude of $\mathrm{1.2\times10^{-4}\text{ }m^3/m^3}$ after 30 iterations. In this study, 30 iterations of the clustering algorithm are used.

In the second step, $\mathrm{N}$ models, $\hat{f}_1, \hat{f}_2,\dots,\hat{f}_N$ are developed using LST, 3-day PPT, LAI, LC, SM at 1 km and SM at 10 km as inputs to the regularized kernel regression algorithm described in Section~\ref{Sec:OurModel}, using training set. \textcolor{black}{The training set was consisted of randomly selected 33\% of the pixels or 500 out of the 2500 pixels that make up the region.} The remaining pixels were used as the test set.  The hard membership of each pixel, $i$, for model development purposes is determined by the maximum value in its membership vector, $\mathbf{m}^i=(m_1^i,m_2^i,\dots,m_N^i)$. The disaggregated value of SM is computed for each point in the test, represented as a vector, $\mathbf{x'}_i = (\mathrm{LST}_i^\mathrm{1 km}, \mathrm{PPT}_i^\mathrm{1 km}, \mathrm{LAI}_i^\mathrm{1 km}, \mathrm{LC}_i^\mathrm{1 km}, \mathrm{SM}_i^\mathrm{10 km})$ by,

\begin{equation}
\mathrm{SM}_i^\mathrm{1 km} = \mathbf{m}^\mathrm{T}\cdot \left(\hat{f}_1(\mathbf{x'}_i), \hat{f}_2(\mathbf{x'}_i),\dots,\hat{f}_N(\mathbf{x'}_i)\right)
\end{equation}

The SRRM is evaluated using the RMSE and standard deviation of the errors over the entire season. The RMSE over for the entire time-period is assessed for each land-cover. Moreover, the disaggregated SM is compared with the true SM. To evaluate how close the density function of the disaggregated estimates is to the density function of the true SM, the Kulback Liebler-Divergence (KLD) between the density of the estimated observations and the true SM is calculated for different LC's over the season. The KLD is a member of the class of well known f-divergences that convey distances in probability space. Any other f-divergence like the Hellinger distance or $\chi^2$-distance can also be used. \textcolor{black}{A sensitivity analysis was conducted to determine how each auxiliary variable separately contributed to errors in downscaled soil moisture (SM). A single auxiliary variable, LST, LAI or PPT, was allowed to vary for each land cover, while the others were set to their mean values. The relative root mean square errors ($RMSE_r$), $RMSE_r=\frac{\bigtriangleup RMSE}{RMSE_{org}}$, were investigated, where $\bigtriangleup RMSE$ is the change in RMSE when a single auxiliary variable is used compared to when all the auxiliary variables are used ($RMSE_{org}$) for each day in 2007. The daily $RMSE_r$ averaged over each LC, baresoil, corn and cotton, in 2007 is also studied.}

In addition, 5 days were selected from the season to understand the effect of the heterogeneity in inputs on the error in disaggregated SM.  Variabilities in precipitation, ranging from uniformly wet to uniformly dry, and in land cover, ranging from bare soil to vegetated with both cotton and sweetcorn, were used as criteria for selecting the days, as shown in Table~\ref{tab:conditions}. Quantitative analyses of spatial variations in SM observed under dynamic vegetation and heterogeneous land cover conditions provide an index of dynamic errors that can be expected. \textcolor{black}{The utility of using multiple models in the region, i.e. one model for each cluster, was also investigated by comparing the disaggregation results to when the entire dataset is considered as a single cluster and only one model is used for disaggregation, on DoY 222 of the study.}

\begin{figure}
\centering\noindent\includegraphics[width=7 in]{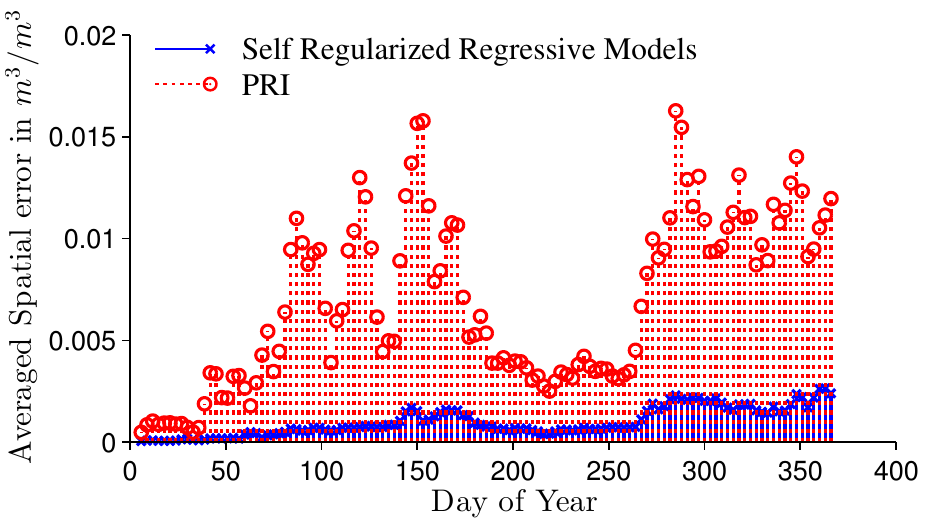}
\caption{Spatially averaged root mean square error in disaggregated Soil Moisture at 1 km for each day of the year in the simulation period using the SRRM and PRI method. }
\label{fig:RMSE}
\end{figure}

\begin{figure}
\centering\noindent\includegraphics[width=3.5 in]{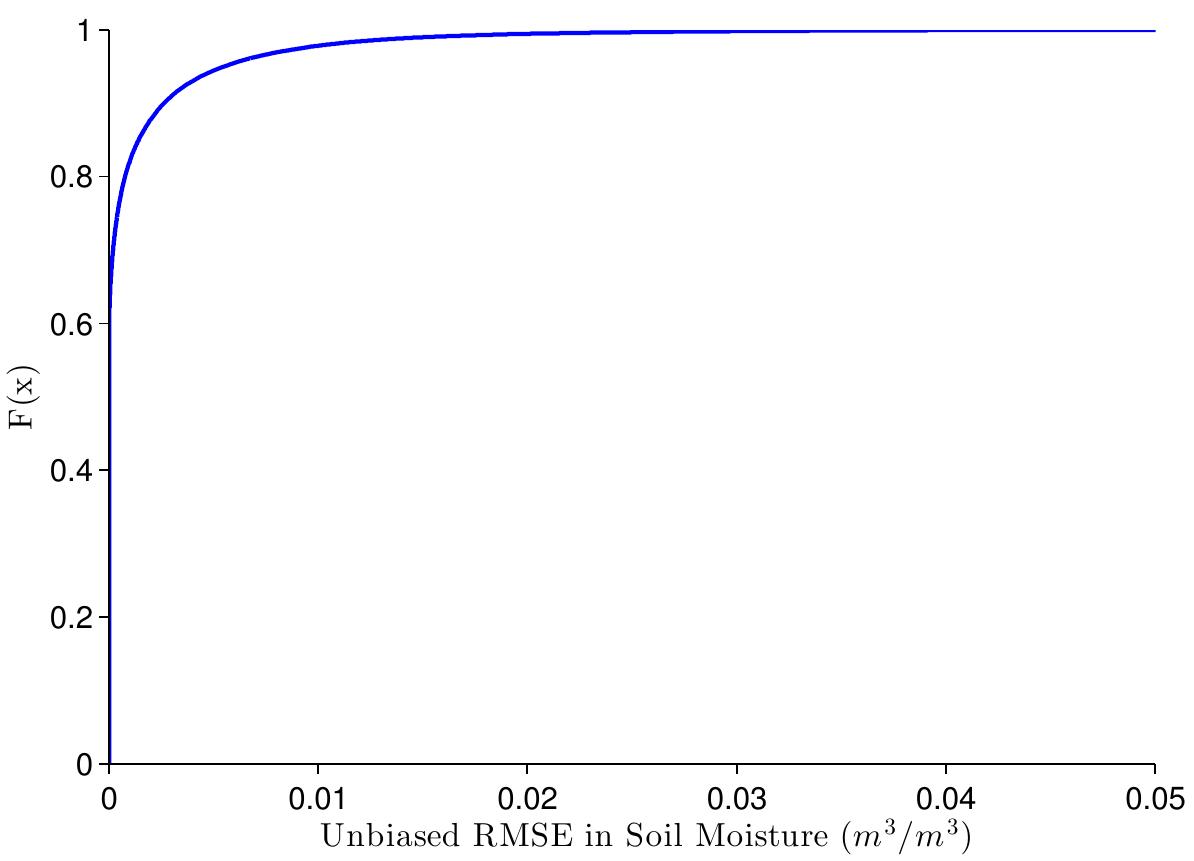}
\caption{Cumulative density function of the errors in soil moisture.}
\label{fig:cdf}
\end{figure}

\begin{figure}
\centering\noindent\includegraphics[width=7 in]{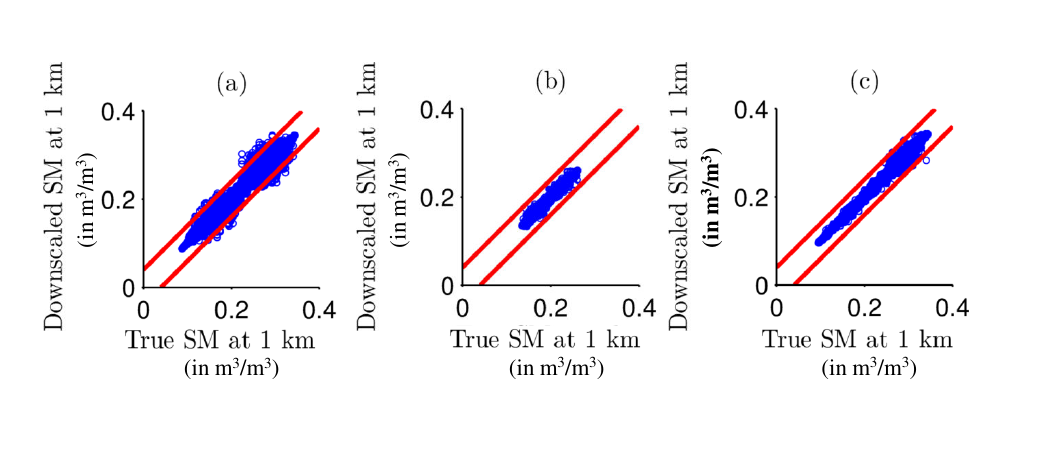}
\caption{Disaggregated Soil Moisture vs. True Soil Moisture at 1 km during the whole season for (a)baresoil pixels (b)corn pixels, and (c)cotton pixels. Lines corresponding to 4\% soil-moisture are shown for each plot.}
\label{fig:scatterlc}
\end{figure}

\begin{figure}
\centering\noindent\includegraphics[width=7 in, angle=90]{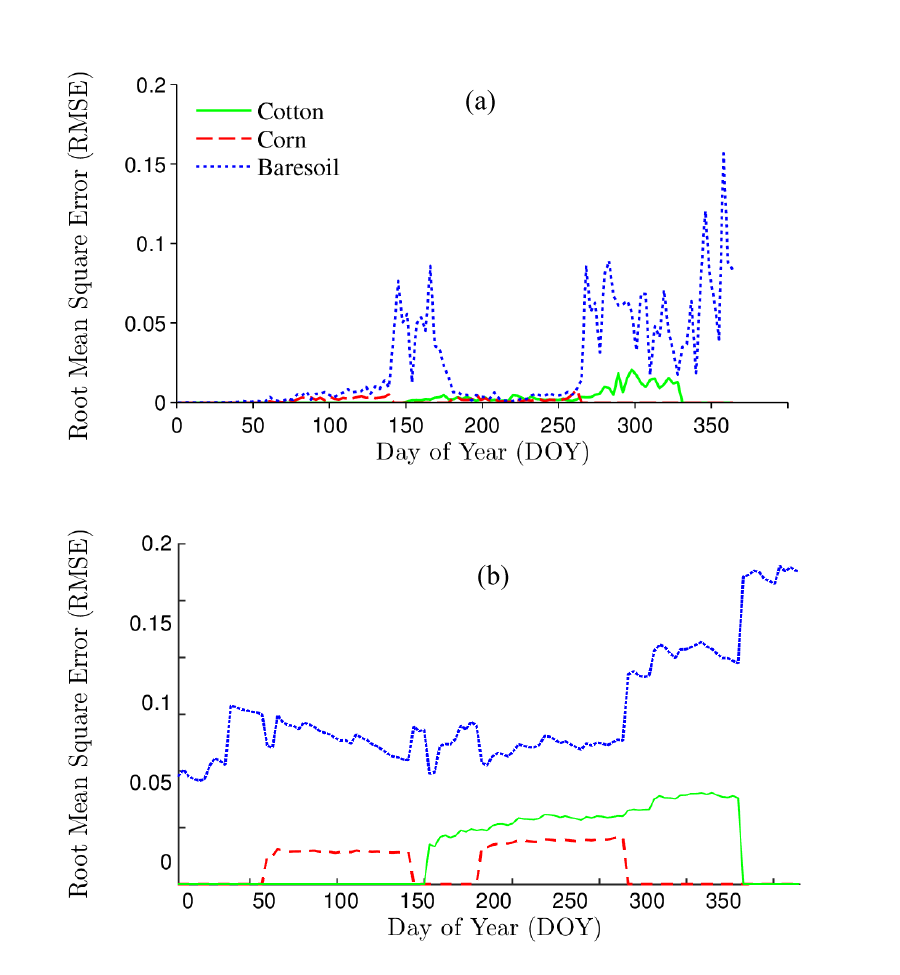}
\caption{Spatially averaged root mean square error in disaggregated Soil Moisture at 1 km for each day of the year in the simulation period for baresoil, corn and cotton landcovers using the (a) SRRM and (b) PRI method.}
\label{fig:errorlc}
\end{figure}

The spatially averaged RMSE for each DoY in the simulation period is shown in Figure~\ref{fig:RMSE}. A Z-test was performed to evaluate whether the disaggregated SM at 1~km is within a standard deviation of $\pm 0.04$ $\mathrm{m^3/m^3}$ from the true SM at 1~km, for meaningful use in hydrological models\cite{das2011}. This null hypothesis was found to be true for every day of the simulation period. Figure~\ref{fig:cdf} shows the cumulative density function (CDF) of the errors in disaggregated SM. About 98\% of the days have an RMSE of less than $0.02$ $\mathrm{m^3/m^3}$ in the disaggregated SM. Figure~\ref{fig:scatterlc} shows the disaggregated SM versus true SM at 1~km. \textcolor{black}{The algorithm does not introduce any bias and the data points are scattered around the $\hat{y}-y=0$ line, with a positive variance.  Most of the points for sweet-corn pixels and all of the points for cotton lie within $0.04$ $\mathrm{m^3/m^3} $.} Figure~\ref{fig:errorlc}(a) shows the errors for each DoY segregated by type of LC. Baresoil pixels during periods of vegetation have the highest RMSE. This is due to sub-pixel vegetation at 250~m within a pixel classified as a baresoil pixel, when the vegetation fraction is $<$ 0.5 at 1~km. Table~\ref{tab:pritable} shows the KLD between the densities of the disaggregated estimates and the true SM. Baresoil pixels at 1~km without any vegetation at 250~m have the lowest KLD. Baresoil pixels at the end of the season, that are affected by remnant crops and baresoil pixels at 1~km with partial vegetation cover at 250~m, have a higher KLD, but very close to 0. Vegetated pixels at 1~km contain a higher KLD as well. The boundary pixels classified as bare-soil have vegetation at the 250 m scale contributing to these errors. 

\begin{figure}
\centering\noindent\includegraphics[width=7 in]{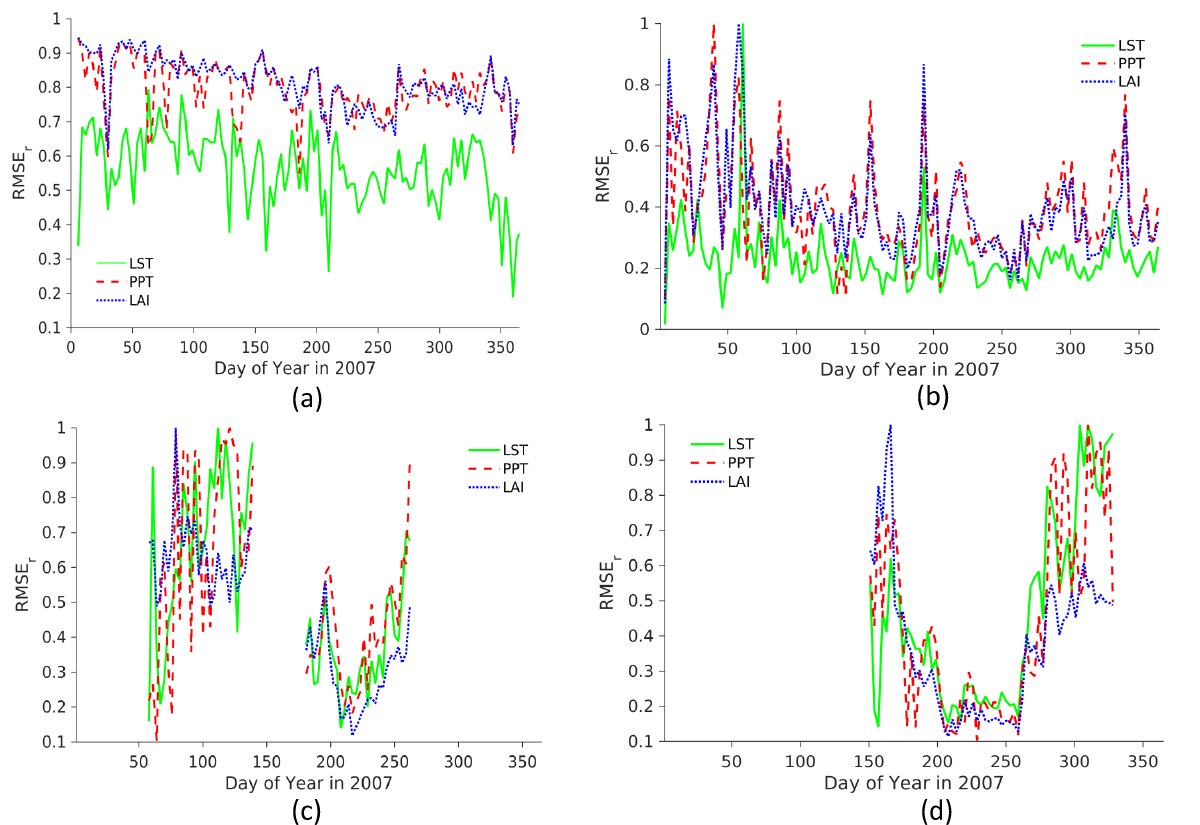}
\caption{(a) Relative change in RMSE ($\mathrm{RMSE_r}$) when only LST, LAI or PPT is used as input for disaggregation in (a) the whole region, (b) bare soil, (c) corn, and (d) cotton for each day of 2007.}
\label{fig:sensitivity}
\end{figure}

\textcolor{black}{Among the three scenarios considered for the sensitivity analysis, RMSEs in downscaled SM are the lowest when just LST is used for disaggregation. This suggests that SM is more strongly coupled to LST than LAI or PPT. This is expected since the spatial patterns apparent in LST images also appear in the SM image, especially baresoil pixels, as shown in Figure \ref{fig:sensitivity}(b), with a weaker and more complex relationship in corn and cotton, as shown in Figures \ref{fig:sensitivity}(c) and (d) respectively. LAI shows higher and similar effects on errors in disaggregated SM for during the mid and late growing seasons of corn and cotton crops.  The use of PPT to disaggregate SM results in a lower RMSEs immediately following a major rainfall event. At other times, its sensitivity to SM is comparable to LST, for baresoil pixels, and to LAI, for vegetated pixels.}

\begin{figure}
\centering\noindent\includegraphics[width=\textwidth]{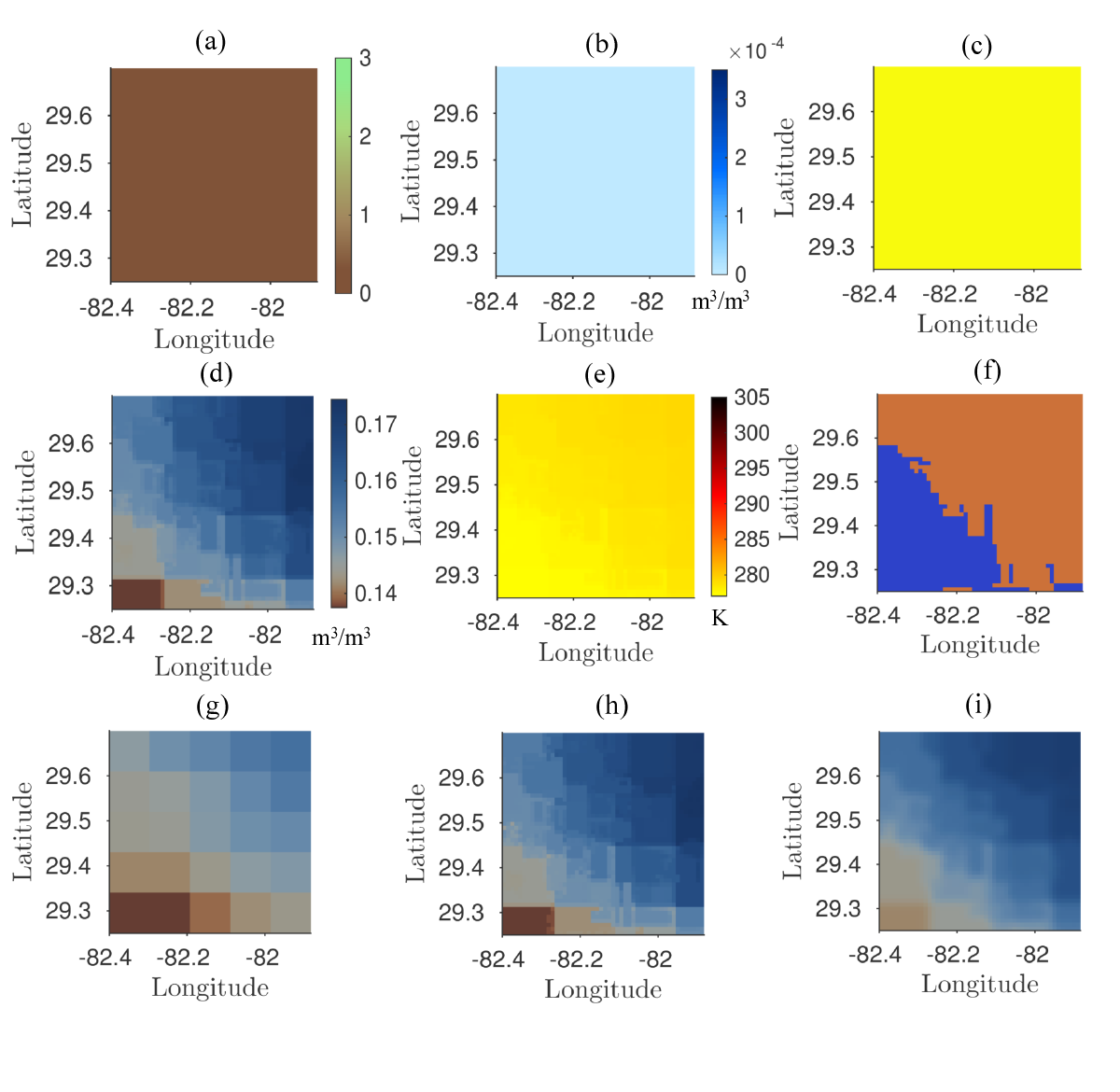}
\caption{DoY 39 - (a) LAI at 1 km, (b) PPT at 1 km, (c) LC at 1 km (yellow represents baresoil), (d) true SM at 1 km, (e) LST at 1 km, (f) clustering result at 1 km, (g) SM observations at 10 km, (h) disaggregated SM using SRRM, (i) disaggregated SM using PRI method.}
\label{fig:39}
\end{figure}

\begin{figure}
\centering\noindent\includegraphics[width=\textwidth]{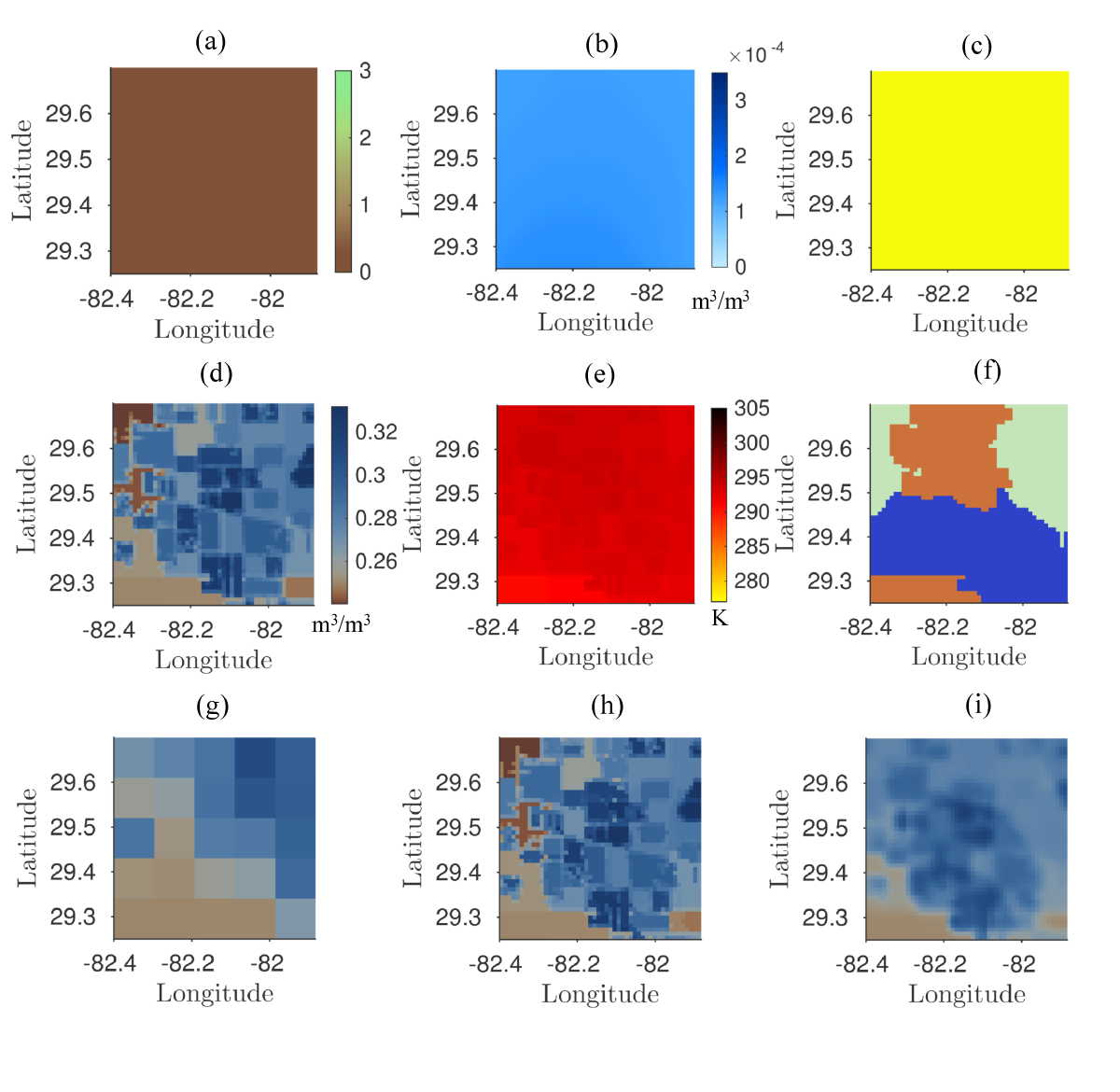}
\caption{DoY 354 - (a) LAI at 1 km, (b) PPT at 1 km, (c) LC at 1 km (yellow represents baresoil), (d) true SM at 1 km, (e) LST at 1 km, (f) clustering result at 1 km, (g) SM observations at 10 km, (h) disaggregated SM using SRRM, (i) disaggregated SM using PRI method.}
\label{fig:354}
\end{figure}

\begin{figure}
\centering\noindent\includegraphics[width=\textwidth]{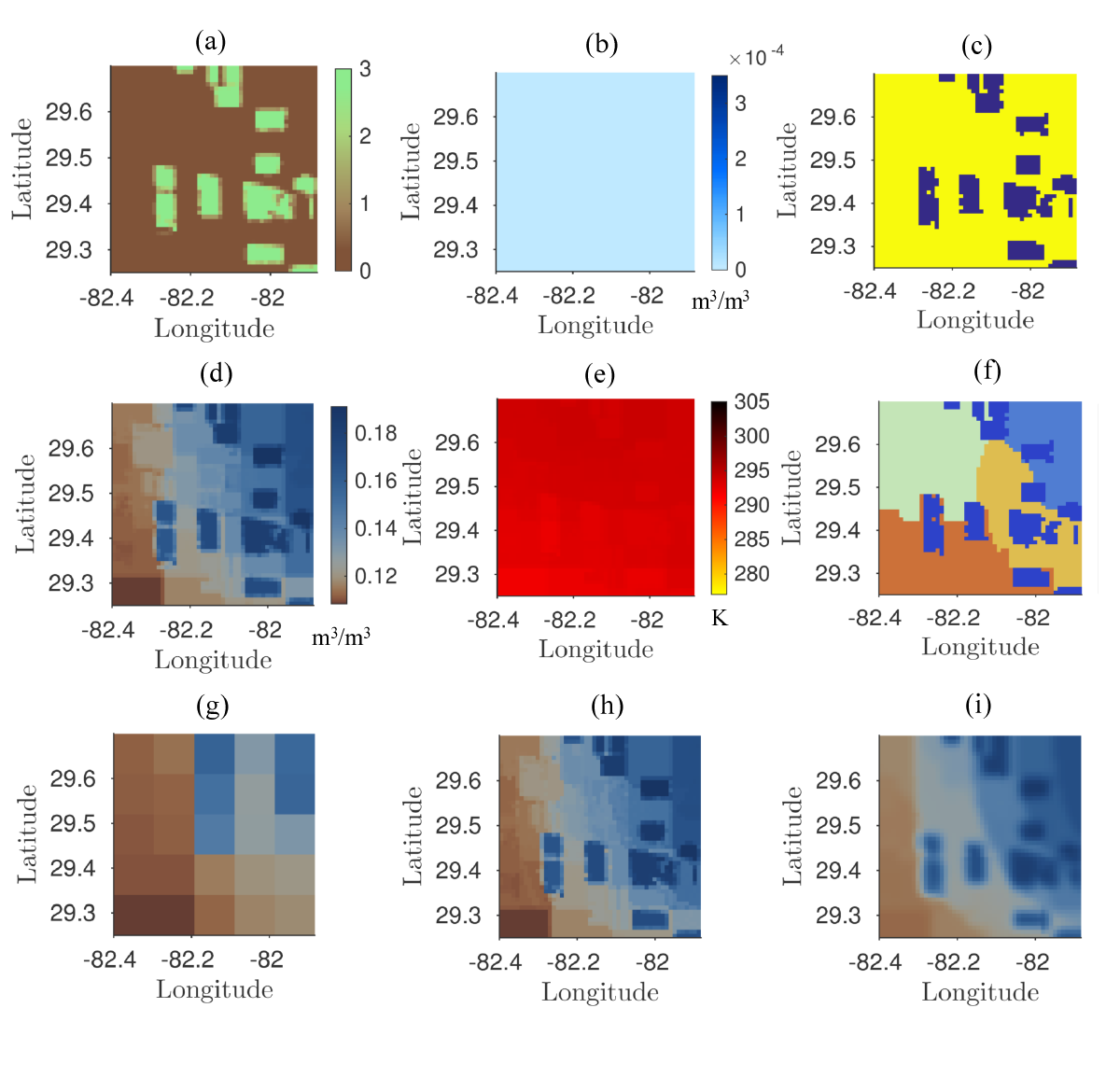}
\caption{DoY 135 - (a) LAI at 1 km, (b) PPT at 1 km, (c) LC at 1 km (yellow represents baresoil, blue represents corn), (d) true SM at 1 km, (e) LST at 1 km, (f) clustering result at 1 km, (g) SM observations at 10 km, (h) disaggregated SM using SRRM, (i) disaggregated SM using PRI method.}
\label{fig:117}
\end{figure}

\begin{figure}
\centering\noindent\includegraphics[width=\textwidth]{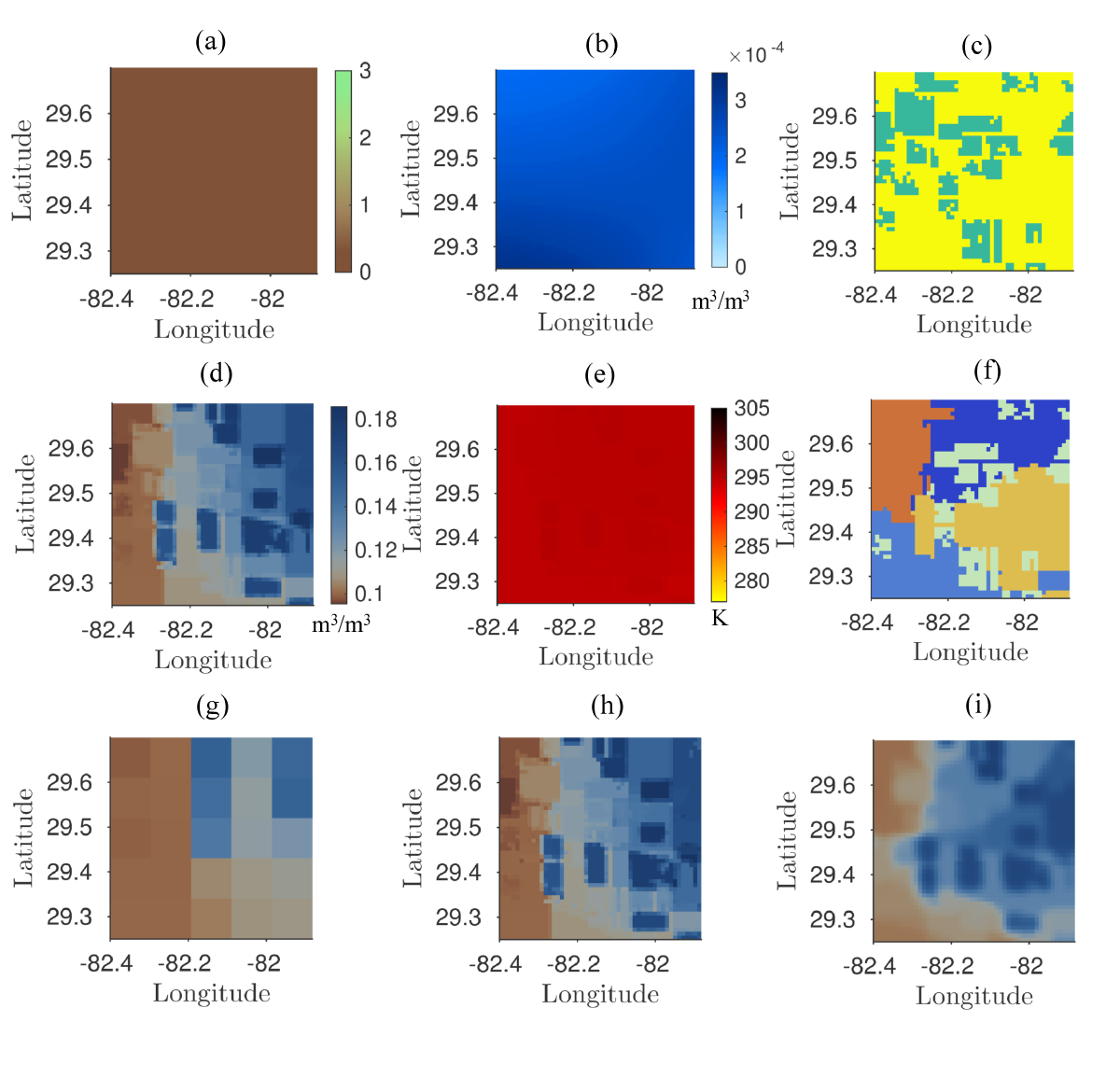}
\caption{DoY 156 - (a) LAI at 1 km, (b) PPT at 1 km, (c) LC at 1 km (yellow represents baresoil, green represents cotton), (d) true SM at 1 km, (e) LST at 1 km, (f) clustering result at 1 km, (g) SM observations at 10 km, (h) disaggregated SM using SRRM, (i) disaggregated SM using PRI method.}
\label{fig:156}
\end{figure}

\begin{figure}
\centering\noindent\includegraphics[width=\textwidth]{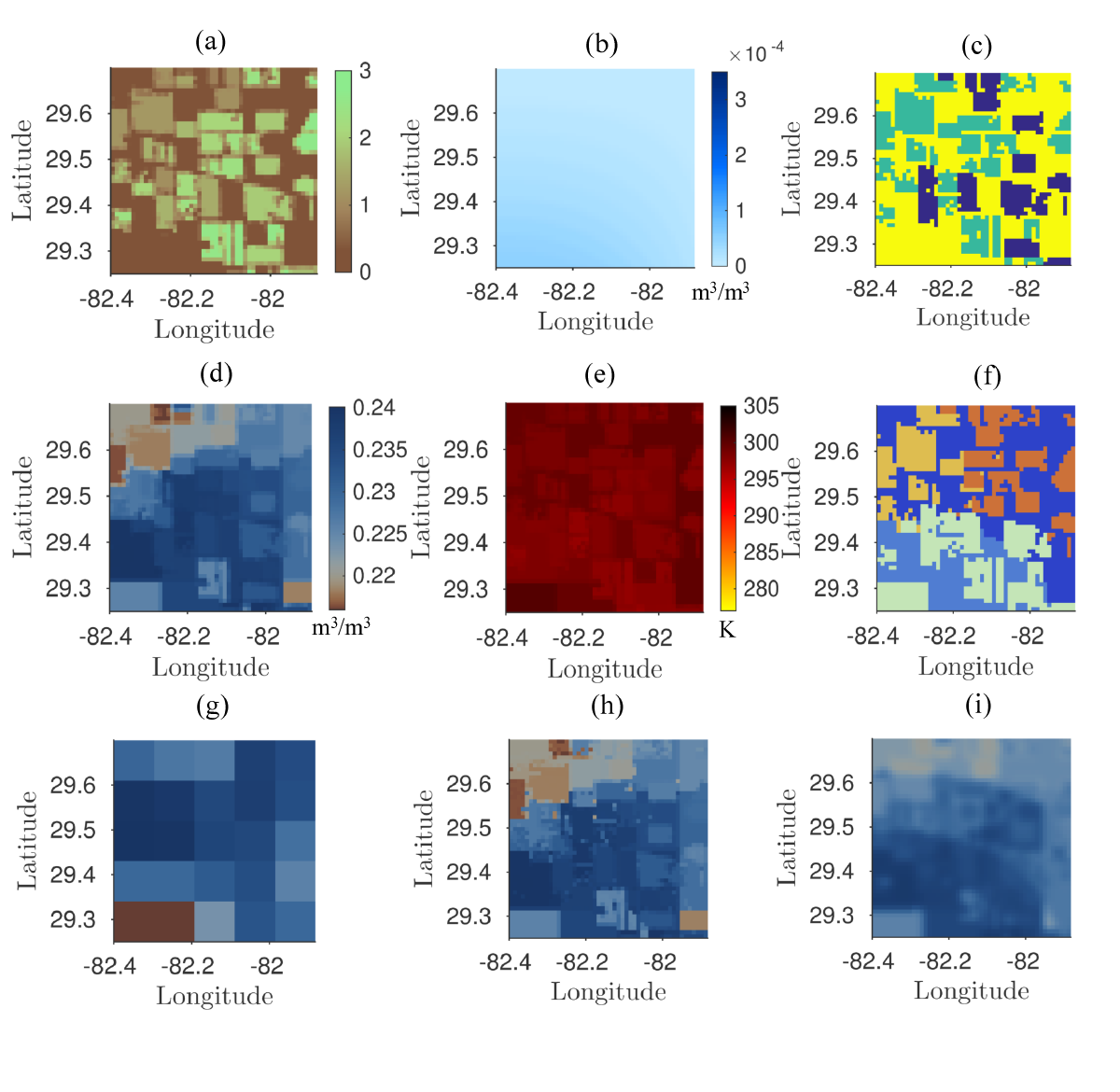}
\caption{DoY 222 - (a) LAI at 1 km, (b) PPT at 1 km, (c) LC at 1 km (yellow represents baresoil, blue represents corn and green represents cotton), (d) true SM at 1 km, (e) LST at 1 km, (f) clustering result at 1 km, (g) SM observations at 10 km, (h) disaggregated SM using SRRM method, (i) disaggregated SM using PRI method.}
\label{fig:222}
\end{figure}

\begin{figure}
\centering\noindent\includegraphics[width=0.5\textwidth,trim={5cm 0 0 0},clip]{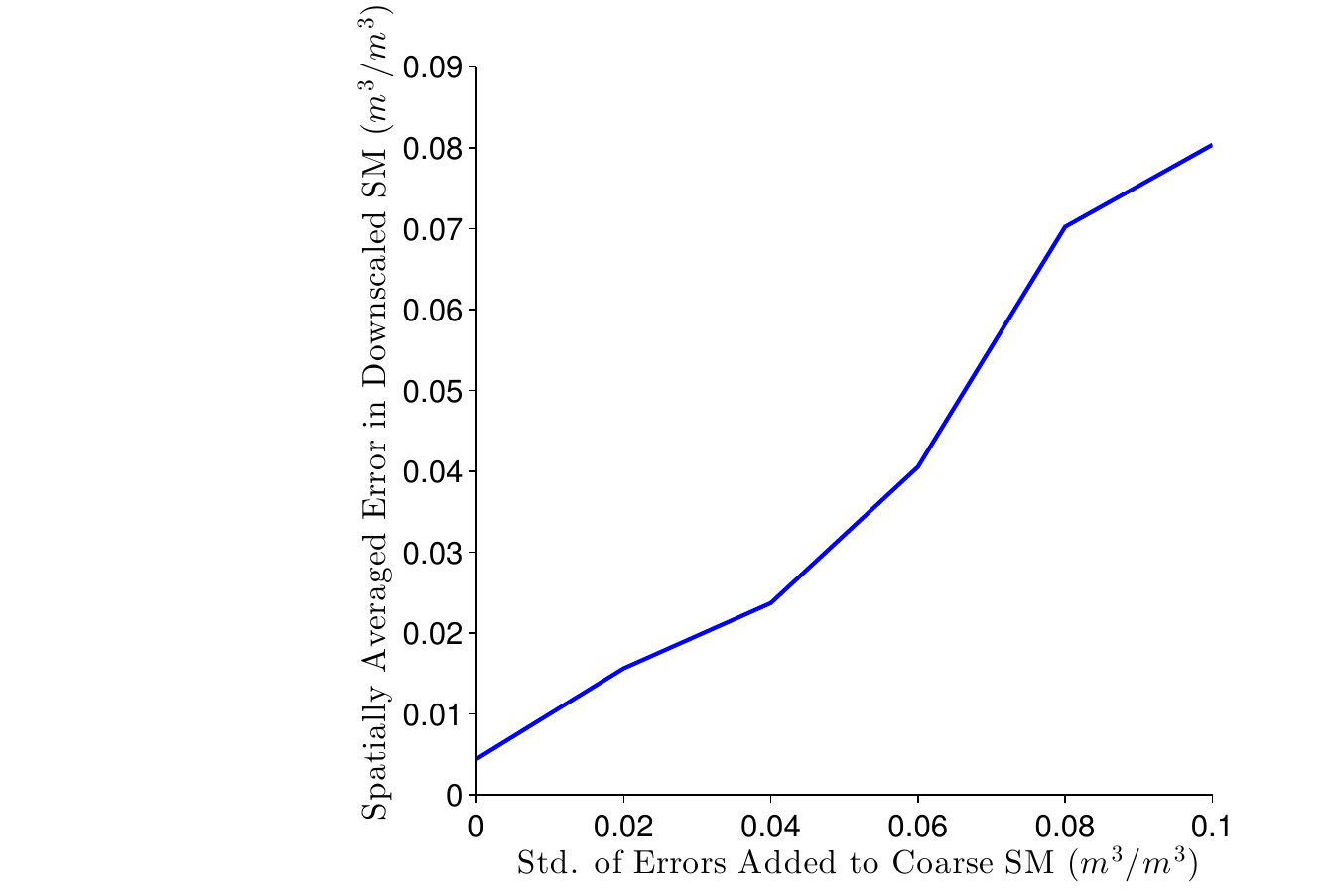}
\caption{Standard deviation of noise added to coarse scale SM vs. unbiased RMSE in disaggregated SM}
\label{fig:prop}
\end{figure}

For the five selected days, the inputs, \textcolor{black}{clustering} results, the first SM estimate, and PRI disaggregated SM are shown in Figures~\ref{fig:39}-\ref{fig:156}. \textcolor{black}{The clustering results indicate that the implicit inclusion of spatial coordinate information adequately constrains the clusters from becoming too small, while the LST, LAI, PPT and LC ensure that the clusters are simultaneously representative of the land-cover and meteorological conditions in the region.  When fields are significantly smaller than the resolution of auxiliary variables, for example, in developing countries, the implicit inclusion of coordinates might not result in a clustering that accurately follows field boundaries, although it would still separate out regions with different meteorological conditions. This would reduce the accuracy of disaggregated SM at the field edges may be reduced and post-processing based on finer scale land-cover will be needed in such scenarios.} Both DoY 39, shown in Figure~\ref{fig:39} and DoY 354, shown in Figure \ref{fig:354} are during  bare soil land cover before and after the growing seasons, respectively. The disaggregated estimates for both days are very close to the true SM at 1 km, but due to crop residue and slightly heterogeneous precipitation in the region (Figure~\ref{fig:354}b), the error for DoY 354 is higher than for DoY 39. It was found that heterogeneity in any one input, is enough to capture vegetation patterns in the disaggregated estimate using Kernel regressive models as shown in Figures \ref{fig:354}a, and ~\ref{fig:117}a, for corn and cotton, when the LST is fairly uniform across the region, while PPT is heterogeneous due to precipitation patterns. On DoY 222, even when there was maximum heterogeneity in LC with corn, cotton, and bare soil, the error in SM is minimal as shown in Figure~\ref{fig:222}. \textcolor{black}{The effects of noise amplitude in the coarse scale SM on the disaggregated SM were also investigated on DoY 222. Independent Gaussian noise with zero mean and standard deviations ranging from 0 to 0.1 $\mathrm{m^3/m^3}$ was added to the coarse-scale SM and the spatially averaged unbiased RMSE in disaggregated SM is shown in Figure~\ref{fig:prop}. The errors grow sub-linearly, i.e. with a slope lower than 1, while the uncertainty in SM is $<$ 0.06 $\mathrm{m^3/m^3}$. When the uncertainty in coarse SM is $>$0.06 $\mathrm{m^3/m^3}$ the errors grow with a slope of 1.14 showing that the errors in the disaggregated SM have a higher magnitude than the uncertainties added to coarse SM. }
 
\begin{figure}
\centering\noindent\includegraphics[width=\textwidth,trim={0cm 0 0 0},clip]{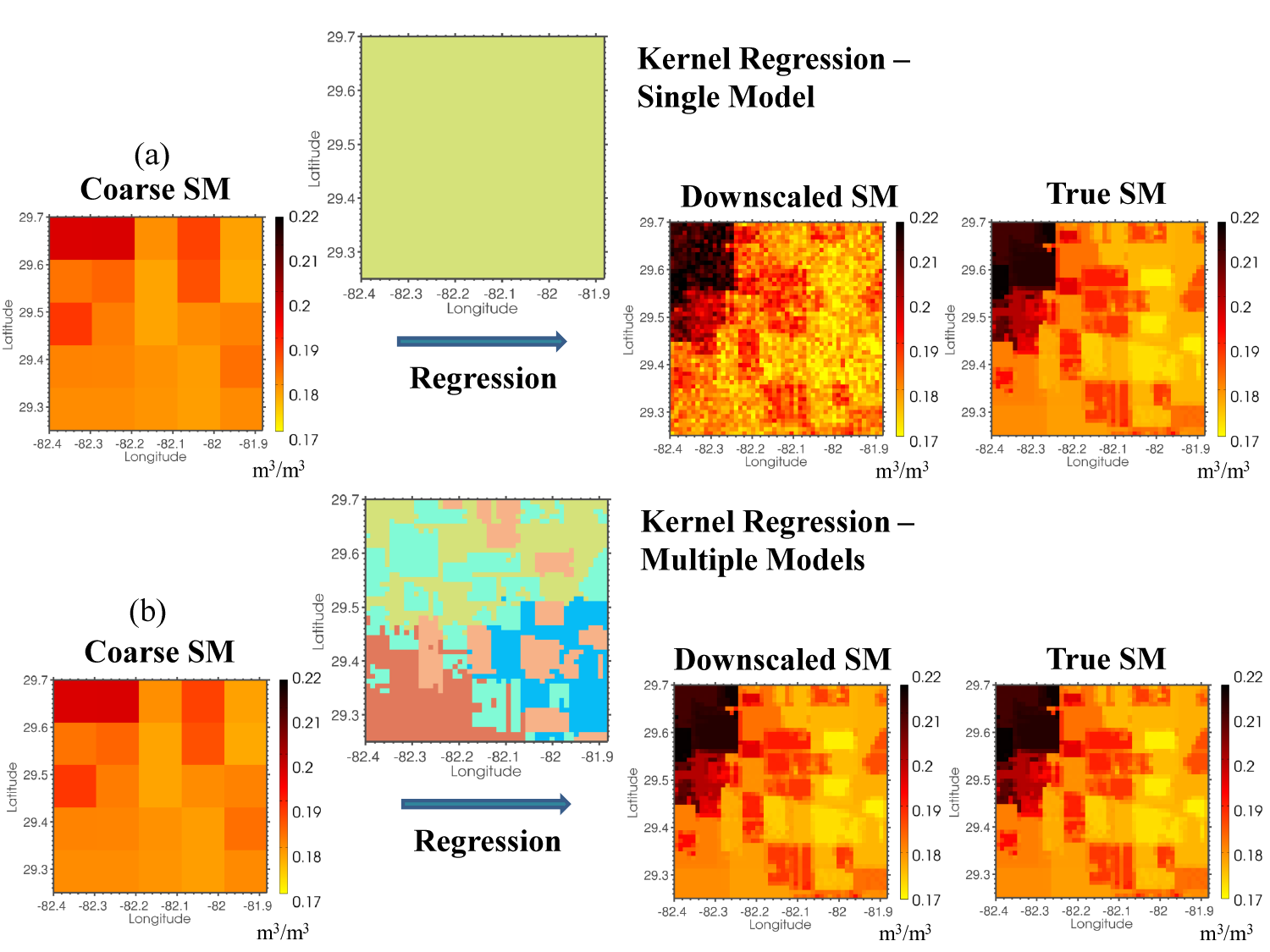}
\caption{SM at 10 km, true SM at 1km and disaggregated SM at 1km using (a) a single cluster for the study region, and (b) multiple clusters following the SRRM algorithm.}
\label{fig:modelnumber}
\end{figure}

\textcolor{black}{The novelty and efficacy of this disaggregation algorithm lies in the utilization of multiple models using clusters. TAs evident in Figure~\ref{fig:modelnumber}(a), a regression model based on a single cluster fails to fit the coarse SM and auxiliary data with a sufficient degree of accuracy, resulting in speckle noise in the disaggregated soil moisture. Instead, Figure~\ref{fig:modelnumber}(b) shows that using multiple cluster-based models is an elegant solution that adequately fits the coarse SM and the auxiliary data, and provides disaggregated estimates of SM with low RMSE.}

\subsection{Comparison between SRRM \& PRI}

The PRI method uses LST, 3-day PPT, LAI, LC, and SM at 1 km every 3 days as input to obtain the first estimate of the SM. To disaggregate SM, in Equation~\ref{eq:bayes}, $\mathbf{X}$ is set to $\{\mathrm{LST},\mathrm{PPT},\mathrm{LAI},\mathrm{LC}\}$ and $\mathbf{y_{\mathrm{train}}}$ is set to $\{\mathrm{SM}_{\mathit{insitu}}\}$. In this study, 33\% of the data set, \textcolor{black}{selected randomly}, is used for training the parametric Bayesian model. For the second step, in Equation~\ref{eq:pri}, the SM observations at 10 km are set as $\mathbf{y}_{\mathrm{coarse}}$ and first estimates of SM at 1 km from the transformation function are set as $\mathbf{y}_{\mathrm{initial}}$. The value of $\mathbf{m}$ after the cost function $\mathrm{I}(\mathbf{m})$ is minimized is the disaggregated SM estimates.

The disaggregated estimates using the SRRM were compared with the PRI estimates using the RMSE and the KLD of the estimated densities of the disaggregated observations. The RMSE over for the entire time-period is assessed for each land-cover using the SRRM and PRI algorithms are compared. The spatial errors are also compared for the selected five days during the simulation period, representing different micro-meteorological and land cover conditions. Finally, the running time of the SRRM and PRI algorithm are compared to understand the effects of the difference in algorithm complexity of the two algorithms.

Figure~\ref{fig:RMSE} shows that the RMSE of the disaggregated observations using the self-regularized regressive models was less than the RMSE using the PRI algorithm. \textcolor{black}{The trends observed when the PRI algorithm is used, such as higher errors during periods of vegetation, are preserved when SRRM is used. However, variations in the difference observed between the SRRM and PRI based SM, can be explained by its different correlations to LC and micro-meteorological conditions. The use of separate models in the SRRM enables low RMSEs even under highly heterogeneous LC. In contrast, the RMSEs increase by a larger magnitude during heterogeneous LC periods for the PRI algorithm because it uses a single disaggregation model for the whole study region.}

Table~\ref{tab:pritable} compares the KLD between the disaggregated estimates generated by the SRRM and PRI algorithms, and true SM at 1~km. \textcolor{black}{The general trends of KLD over different LC conditions followed by the SRRM are similar to those observed for the PRI. However, the errors for each LC are individually lower for the SRRM compared to the PRI method, as shown in Figure~\ref{fig:errorlc}. This is further validated by the KLD for the SRRM estimates that is 3 orders of magnitude less than for PRI estimates. }

Figures~\ref{fig:39}-\ref{fig:156} compare the disaggregated SM estimates using the SRRM to those using the PRI method. The estimates using PRI does not have sharply defined regions, unlike those observed in the disaggregated SM using SRRM. \textcolor{black}{The sharpness of disaggregated result could arise either from noise or spatial discontinuities in the inputs due to physical discontinuities in meteorological or land-cover conditions. Any disaggregation algorithm must maintain the latter, while suppressing the former. The equation~\ref{eq:pri}, with $\beta = 2$ maximizes a cost-function that blurs the disaggregated SM so that the median error over all pixels is minimized at the cost of a greater variance in error. In the SRRM, the use of multiple models based on clusters ensures that the spatial discontinuity is maintained in the disaggregated SM when it is caused by a physical discontinuity. If the discontinuity originates from additive noise, under certain assumptions, the kernel regression suppresses the discontinuity in the disaggregated SM. The assumptions are that the noise is spatially un-correlated and has a wide probability density function. The results in ~\cite{sabater2014} show that both assumptions are reasonable. } 

\textcolor{black}{The average execution time of the PRI was about 1.56 hours/disaggregation day and that of the SRRM was about 32 minutes/disaggregation day. This is expected because the complexity of PRI is $\mathcal{O}(N^3\prod k_j)$  where $k_j$ are the number of bins used to estimate the PDF. For an adequate estimate of the PDFs, $\prod k_j\approx N$ and the complexity approaches $\mathcal{O}(N^4)$, that is an order of magnitude higher than the complexity of the SRRM based algorithm.}

Thus, the SRRM achieves low mean errors using a non-linear regression and low error variances using multiple regressive models with soft boundaries. This ensures that sharpness is maintained along with low RMSEs. Given exhaustive training data, the PRI algorithm will have similar performance as the SRRM as shown in~\cite{Chakrabarti2014}. However, for operational use of the methodologies at field-scale in regions with highly varied LC or micro meteorology with low volume of training data. SRRM provides sharp images of disaggregated SM \textcolor{black}{ with a faster run time}, less complexity, and lower RMSEs compared to the PRI. 
\section{Conclusion}
\label{Sec:Conclusion}

In this study, a disaggregation methodology based upon SRRM was developed, implemented and evaluated that preserves the high variability in SM due to heterogeneous meteorological and vegetation conditions. The SRRM preserves heterogeneity by utilizing a clustering algorithm to create a number of regions of similarity which subsequently, are used in a kernel regression framework. The clusters were computed using RS products, \textit{viz.} PPT,  LST, LAI, and LC. The kernel regression was implemented on the clusters using \emph{in-situ} SM. 96\% of the pixels across the whole season were found to have a disaggregation error of less than $0.02$ $\mathrm{m^3/m^3}$. The KLD values for disaggregated SM at 1 km for the SRRM  was equal to 0, for all land covers. In contrast, the PRI method has KLD values that are several orders of magnitude higher, and has an threefold higher execution time. The averaged spatial error is also markedly lower for the SRRM compared to the PRI method.

It is envisioned that the SRRM will be implemented and evaluated in this study may be applied using satellite images. For example, the PPT data may be obtained from the Global Precipitation Measurement missions and the LAI, LST and LC products are available from the MODIS sensor aboard Aqua and Terra satellites.

\renewcommand{\baselinestretch}{1.0}
\bibliographystyle{IEEEtran}
\bibliography{combined}

%
\clearpage
\label{Tables}
\listoftables
\clearpage
\begin{table}
\renewcommand{\arraystretch}{1.5} 
\centering
\caption{Planting and harvest dates for sweet corn and cotton during the 2007 growing season}
\label{tab:crops}
\begin{tabular}{c|c|c}
\hline
\multicolumn{1}{c|}{Crop} & \multicolumn{1}{c|}{Planting DoY} & \multicolumn{1}{c}{Harvest DoY} \\
\hline
Sweet Corn & 61 & 139 \\
 & 183 & 261 \\
Cotton & 153 & 332 \\
\hline        
\end{tabular}
\end{table}

\clearpage
\begin{table}
\renewcommand{\arraystretch}{1.5} 
\centering
\caption{Days selected for evaluating SRRM estimates. These days capture variability in precipitation/irrigation (PPT) and land cover (LC) }
\label{tab:conditions}
\begin{tabular}{r|l|l}                
\hline
\multicolumn{1}{r|}{DoY} & \multicolumn{1}{c|}{PPT} & \multicolumn{1}{c}{LC} \\
\hline
39 & Dry & Bare \\
135 & Dry, Irrigated & Sweet Corn \\
156 & Wet & Cotton \\
222 & Dry, Irrigated & Sweet Corn and Cotton \\
354 & Wet & Bare \\
\hline
\end{tabular}
\end{table}

\clearpage
\begin{table}
\renewcommand{\arraystretch}{1.5} 
\centering
\caption{KL divergence over the 50$\times$50 km$^2$ region for the disaggregated estimates of SM obtained at 1 km using the SRRM and PRI method.}
\label{tab:pritable}
\begin{tabular}{c|c|c}
\hline
\multicolumn{1}{c|}{Land Cover} & $\mathrm{KLD_{SRRM}}$  & $\mathrm{KLD_{PRI}}$\\
\hline

Corn & $1.8615\times 10^{-17}$ & 0.0234\\
Cotton &$2.4828\times 10^{-04}$ & 0.0283 \\
Baresoil\footnote{Baresoil pixels with vegetated sub-pixels at 250 m till DoY 332} & $5.6222\times 10^{-5}$ & 0.1036\\
Baresoil\footnote{Baresoil pixels after DoY 332}& $5.628\times 10^{-6}$ & 0.0120 \\
Baresoil\footnote{Baresoil pixels without any  vegetated sub-pixels at 250 m till DoY 332} & $2.5948\times 10^{-6}$ & 0.0114 \\
\end{tabular}
\end{table}

%
\clearpage
\label{Figures}
\listoffigures
\clearpage

\clearpage

\clearpage

\clearpage

\clearpage

\clearpage

\clearpage

\clearpage

\clearpage

\clearpage

\clearpage

\clearpage

\clearpage

\clearpage

\clearpage

\clearpage

\clearpage

\clearpage

\end{document}